\documentclass[lettersize,journal]{IEEEtran}
\usepackage{amsmath,amsfonts,bm}
\usepackage{algorithmic}
\usepackage{algorithm}
\usepackage{array}
\usepackage{booktabs}
\usepackage[caption=false,font=normalsize,labelfont=sf,textfont=sf]{subfig}

\usepackage{diagbox}
\usepackage{makecell}
\usepackage{textcomp}
\usepackage{stfloats}
\usepackage{url}
\usepackage{verbatim}
\usepackage{graphicx}
\usepackage{color}
\usepackage{nomencl}
\usepackage{cite}
\usepackage{url}
\usepackage[numbers,sort&compress]{natbib}
\makenomenclature
\begin{document}

\title{Continual Interactive Behavior Learning With Traffic Divergence Measurement: \\A Dynamic Gradient Scenario\\ Memory Approach}

\author{Yunlong Lin, Zirui Li, Cheng Gong, ~\IEEEmembership{Student Member,~IEEE}, Chao Lu, ~\IEEEmembership{Member,~IEEE}, Xinwei Wang,  Jianwei Gong*,~\IEEEmembership{Member,~IEEE}
\thanks{This work is supported by the National Natural Science Foundation of China under Grants 61703041 and U19A2083, and is also supported by China Scholarship Council (CSC). (Yunlong Lin and Zirui Li contribute equally in this work.)}
\thanks{Yunlong Lin, Zirui Li, Cheng Gong, Chao Lu and Jianwei Gong are with the School of Mechanical Engineering, Beijing Institute of Technology, Beijing 100081, China (Email\tt\small: yunlonglin@bit.edu.cn; z.li@bit.edu.cn; chenggong@bit.edu.cn; chaolu@bit.edu.cn;  gongjianwei@bit.edu.cn)}
\thanks{Zirui Li is also with the Department of Transport and Planning, Faculty of Civil Engineering and Geosciences, Delft University of Technology, Stevinweg 1, 2628 CN Delft, The Netherlands}
\thanks{Xinwei Wang is with the School of Engineering and Materials Science, Queen Mary University of London, UK
 (Email: \tt\small xinwei.wang@qmul.ac.uk)}%
\thanks{(Corresponding authors: C. Lu and J. Gong)}}

\maketitle

\begin{abstract}
Developing autonomous vehicles (AVs) helps improve the road safety and traffic efficiency of intelligent transportation systems (ITS). Accurately predicting the trajectories of traffic participants is essential to the decision-making and motion planning of AVs in interactive scenarios. Recently, learning-based trajectory predictors have shown state-of-the-art performance in highway or urban areas. However, most existing learning-based models trained with fixed datasets may perform poorly in continuously changing scenarios. Specifically, they may not perform well in learned scenarios after learning the new one. This phenomenon is called "catastrophic forgetting". Few studies investigate trajectory predictions in continuous scenarios, where catastrophic forgetting may happen. To handle this problem, first, a novel continual learning (CL) approach for vehicle trajectory prediction is proposed in this paper. Then, inspired by brain science, a dynamic memory mechanism is developed by utilizing the measurement of traffic divergence between scenarios, which balances the performance and training efficiency of the proposed CL approach. Finally, datasets collected from different locations are used to design continual training and testing methods in experiments. Experimental results show that the proposed approach achieves consistently high prediction accuracy in continuous scenarios without re-training, which mitigates catastrophic forgetting compared to non-CL approaches. The implementation of the proposed approach is publicly available at \url{https://github.com/BIT-Jack/D-GSM}. 
\end{abstract}


\nomenclature[00]{$\bf{Symbol}$}{$\bf{Definition}$}
\nomenclature[05]{$t_h$}{Observed time horizon}
\nomenclature[06]{$t_f$}{Predicted time horizon}
\nomenclature[07]{$\bf{tr}$}{Positions of vehicles}
\nomenclature[09]{${\bf{\hat tr}}$}{Predicted position of vehicles}
\nomenclature[10]{$N_{ts}$}{Number of testing samples}
\nomenclature[11]{$\bf{X}$}{Historical trajectories observed by predictor}
\nomenclature[12]{$\bf{Y}$}{Future trajectories to be predicted}
\nomenclature[14]{$\bf{\Gamma }$}{Bivariate Gaussian distribution over the predicted time horizon}
\nomenclature[17]{$\bf{{X}}_{cond}^N$}{Condition considered $N$ surrounding vehicles}
\nomenclature[19]{${\rm{CKLD}}_{i,j}^{\rm{wt}}$}{Weighted-CKLD between distribution $p_i$ and $p_j$}
\nomenclature[20]{$m_r^{\rm{D}}$}{Amounts of memory data allocated to the $r^{th}$ task}
\nomenclature[22]{${f_{\bm{\theta}}}$}{Model parameterized by $\bm{\theta}$}
\nomenclature[24]{${f^{'}_{\bm{\theta}}}$}{Model state at the end of learning of the last task}
\nomenclature[26]{$\bf{g}_r$}{Loss gradient of the $r^{th}$ previous task}
\nomenclature[31]{$S$}{Continuous scenarios}
\nomenclature[32]{$d_i$}{The $i^{th}$ scenario in continuous scenarios}
\nomenclature[33]{$c$}{Index of "the current scenario".}

\printnomenclature

\section{Introduction}
\IEEEPARstart{A}{utonomous} vehicles (AVs) play an essential role in improving traffic safety and efficiency in intelligent transportation systems (ITS)~\cite{behavior_ITS}. Since understanding interactive behavior of road users and predicting their future trajectories support the efficient decision-making and enable the risk assessment of AVs, they are fundamental to develop AVs~\cite{traj_review, huang2022differentiable, zhan2018probabilistic}.

In early studies, trajectories of vehicles are predicted based on kinetic and dynamic models~\cite{brannstrom2010model-physics, polychronopoulos2007sensor-physics}. These approaches are classified into physics-based methods in~\cite{lefevre2014survey}. Physics-based methods perform well in short-term (less than 1s) prediction. However, driving behaviors among vehicles influence each other, but these methods do not consider these interactive behaviors. Thus, physics-based methods cannot handle changes caused by the execution of a particular behavior (e.g., a lane-changing action, acceleration, or brake). It limits their long-term prediction performance in interactive scenarios~\cite{lefevre2014survey}.

More advanced interaction-aware methods are developed to overcome these limitations. Most interaction-aware methods are learning-based, which predict future trajectories based on the modeling of interactions between agents~\cite{mozaffari2020deep}. By utilizing Long Short-Term Memory (LSTM), Social-LSTM is the first work that modeled interactions between pedestrians as social behaviors~\cite{alahi2016social-lstm}. In social-LSTM, a grid-based pooling mechanism was designed to represent the spatial information of pedestrians in a scene. Then social-pooling layers were used to capture interactions among pedestrians based on the spatial information.~\cite{deo2018convolutional-lstm} improved the performance of Social-LSTM by applying convolutional layers to replace the fully connected layers.

\begin{figure}[tp]
      \centering
      \includegraphics[scale=1.0]{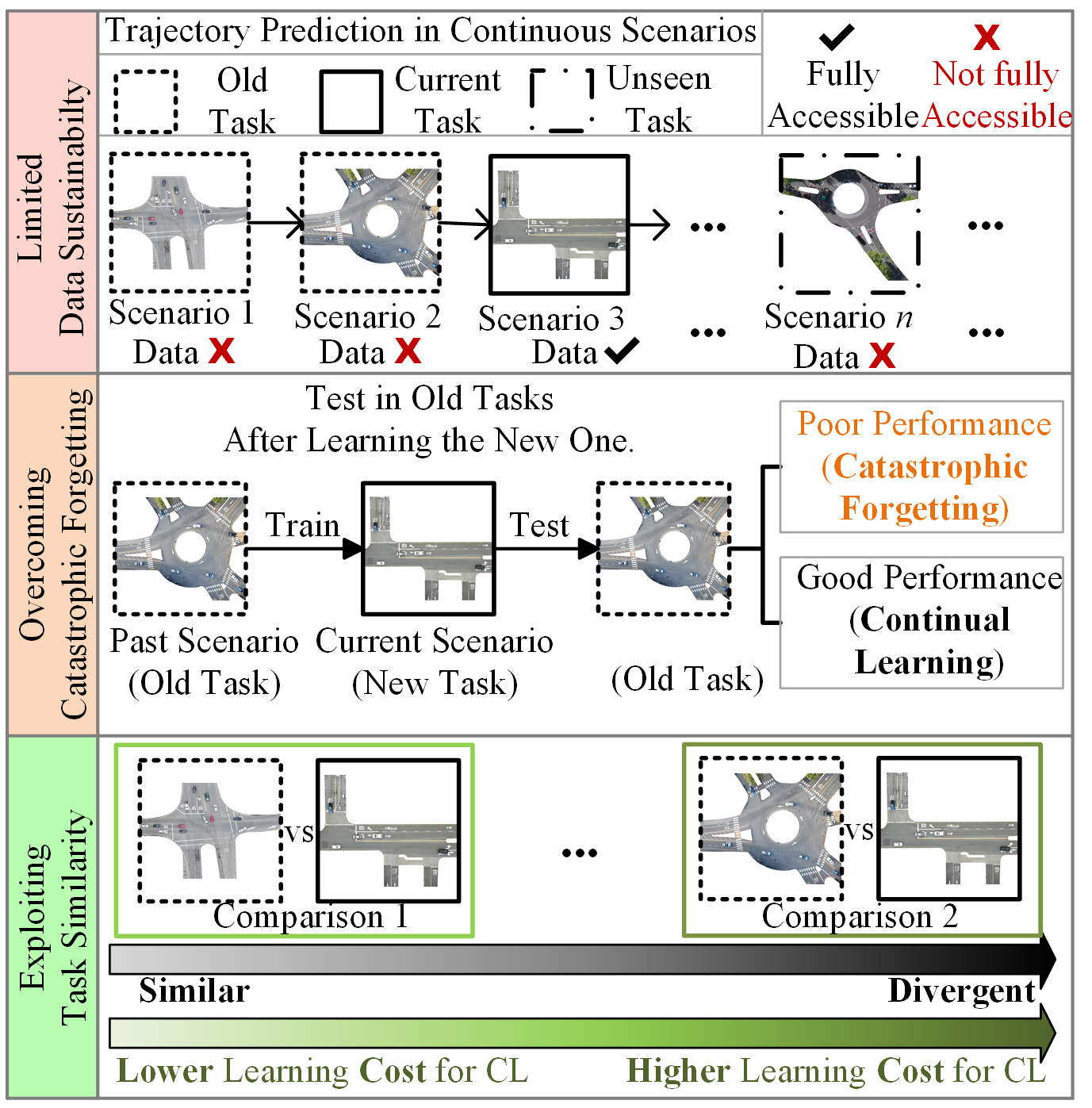}
      \captionsetup{font={small}}
      \caption{Continuous scenarios are coming in a sequence. Due to the limited data sustainability, the data of past and unseen scenarios are not fully accessible. In continuous scenarios, trajectory predictor trained with the current scenario data may have poor performance in learned scenarios (old tasks), which is called catastrophic forgetting. The proposed CL approach aims to address this problem. Measurements of the traffic divergence (task similarity) are also exploited to reduce learning cost in this study.}
      \label{fig_intro}
\end{figure} 

Moreover, Social-LSTM was applied in the generator of Generative Adversarial Networks (GAN) to predict socially plausible trajectories in Social-GAN~\cite{gupta2018socialgan}. Due to the more explicit modeling of interactions, graph-based methods with the better interpretability are developed. Graph-based methods represent interactions between agents by the nodes and edges in graphs. In~\cite{li2021hierarchicalgraph}, interactions and trajectories were modeled as spatial-temporal graphs using a Graph Neural Network (GNN). Then trajectories of vehicles and pedestrians were predicted by a GNN-based multitask learning framework. Similarly,~\cite{sun2020recursive-graph} extracts interactions into social behavior graphs. Then a graph convolutional neural network was applied to propagate social interactions in such graphs, outperforming prior works.

However, the studies above mainly focus on designing networks or architectures to improve the accuracy of predictions. Most existing models are trained and tested specifically for each dataset. Models may fail under this learning paradigm if testing data distribution differs from the training data distribution~\cite{li2020importance,li2022personalized}. Meanwhile, AVs are expected to drive through various scenarios continually in applications. Since factors including traffic rules, the density of traffic flow, and road geometry are divergent, interactive behaviors vary in different scenarios. Most models need to be re-trained on increasing datasets to guarantee the good performance in continuous scenarios~\cite{clrobot_lesort2020continual}. With a heavy computing burden and data storage requirement, this scheme is not efficient and practical. This paper is inspired by Continual Learning (CL)~\cite{chen2018lifelong}, which trains models on new data streams without directly accessing old data to address these problems.

Given a potentially unlimited data stream, learning from a sequence of partial experiences where all data is not available at once is called Continual Learning~\cite{clrobot_lesort2020continual}. In continual learning, the model is updated with a dataset of the current task without directly accessing old ones.
As shown in Fig. \ref{fig_intro}, models in this work observe different scenarios sequentially. Only the data of currently observed scenario are fully available.The data of past scenarios can be stored with limited amounts by CL strategies. Under this assumption, trainable parameters of non-CL models are optimized to minimize the loss of the current scenario. Conversely, the training loss of past scenarios is ignored. Thus, non-CL models may have low predicting accuracy in past scenarios after learning the new one, which is called "catastrophic forgetting". In other words, catastrophic forgetting happens since models try to "fit" the current task without considering the performance on past ones. One of the primary purposes of continual learning strategies is to alleviate catastrophic forgetting~\cite{chen2018lifelong, cossu2021continual_survey2, perkonigg2021dynamic_nature}. To overcome the catastrophic forgetting, this paper aims to enable learning-based trajectory predictors to have consistent good performance in sequential tasks without re-training. The task similarity is also exploited to balance the learning cost and the performance.

More detailedly, this work is inspired by rehearsal methods, which are advanced strategies in CL~\cite{van2020brain-rehearsal, dgr-rehearsal, gem}. Rehearsal methods achieve the goal of CL by using a memory system to keep the learned knowledge. The specific strategy used in this work is Gradient Episodic Memory (GEM)~\cite{gem}. GEM firstly uses the data stored in the episodic memory to calculate losses on previous tasks. Then, these losses are utilized to define an inequality constraint,  interfering with the training process. Under the constraint, models are required to avoid the increment of losses on previous tasks when updating the trainable parameters to mitigate the catastrophic forgetting.

In this research, a novel CL approach termed Dynamic Gradient Scenario Memory (D-GSM) is proposed for the prediction of vehicle trajectory in continuous scenarios. Compared with non-CL models, D-GSM improves the predicting performance in continuous scenarios by constraining the loss increment in observed scenarios. Moreover, inspired by brain science~\cite{kudithipudi2022biological-nature}, a dynamic memory is designed in D-GSM by exploiting the similarity of tasks. The similarity of tasks are modeled by traffic divergence between scenarios. Instead of treating all previous tasks equally as in~\cite{ma2021continual, bao2021lifelong}, the dynamic memory allocates different memory resource to divergent tasks, balancing the performance and training efficiency of the CL strategy in D-GSM. Main contributions of this paper are:
\begin{itemize}
    \item[1)] A novel continual learning approach named Dynamic Gradient Scenario Memory (D-GSM) is proposed for interactive behavior learning in continuous scenarios. With the help of GEM, the proposed D-GSM enables learning-based prediction models to consistently perform well in various scenarios coming in a sequence without re-training by avoiding the increment of losses on previous tasks.
    \item[2)]A new metric based on Kullback-Leibler divergence (KLD) is proposed to measure traffic divergence between different scenarios. The distance between distributions of different scenario data is measured by the KLD-based metric, representing the traffic divergence. Furthermore, a dynamic memory exploiting the measurement of traffic divergence is developed to improve the training efficiency of the proposed CL approach.
    \item[3)]Evaluation methods for vehicle trajectory prediction in continuous scenarios are designed. Then, three experiments are conducted based on the divergent scenario data collected from different locations. It should be noted that the proposed CL approach is a plug-and-play approach. The base model adopted in experiments is demonstrated as an example.
\end{itemize}

\section{Problem Formulation and Preliminaries}
\label{section_formulation}
In this work, interaction-aware predictors are trained using a sequence of datasets collected in different scenarios. To formulate the trajectory prediction in continuous scenarios, the fundamental problem, i.e., the prediction of vehicle trajectory in the interactive scenario, will be firstly formulated in Section \ref{subsection_traj}. Then, based on the fundamental problem, the CL task for trajectory prediction in continuous scenarios will be described in Section \ref{subsection_cl_traj}.

 \begin{figure*}[tp]
      \centering
      \includegraphics[scale=1.0]{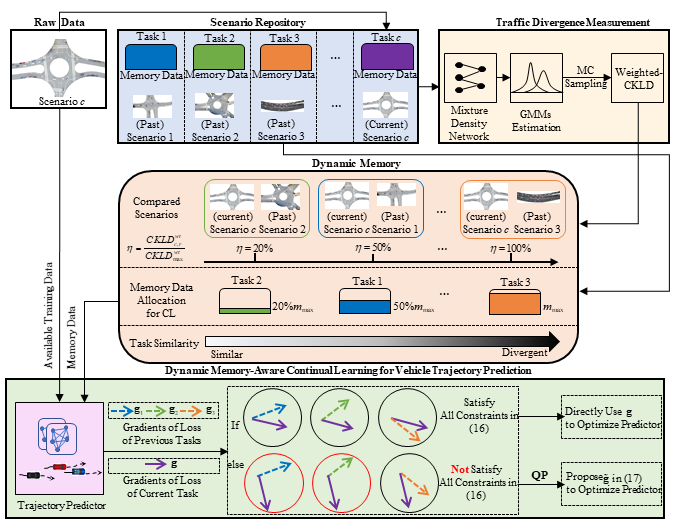}
      \captionsetup{font={small}}
      \caption{Dynamic Gradient Scenario Memory (D-GSM): When a new traffic scenario arrives, scenario repository will firstly store new data as memory data. Then, the traffic divergence measurement module utilizes memory data to measure the divergence between the current scenario and each past scenario. Based on the measuring of divergence, the dynamic memory module allocates memory data with different amounts for different previous tasks to the CL module. Finally, with the help of GEM strategy, the trajectory predictor is trained in the CL module. Since the more memory data bring the higher computing cost, the dynamic memory balances the training efficiency and performance by allocating memory data in a reasonable way.}
      \label{fig_method}
 \end{figure*} 

\subsection{Trajectory Prediction in  Interactive Scenarios}
\label{subsection_traj}
As most interactive behavior learning research~\cite{ li2021interactive-graph, li2022ensemble, li2021hierarchicalgraph}, this work considers scenarios\footnote{The exploration of rigorous definitions of "traffic scenario" or "traffic scene" is not the focus of this work. Researchers who are interested in the exploration of these concepts can refer to~\cite{ulbrich2015defining}.} with highly interactive behaviors among vehicles, including scenarios in urban and highways. The scenarios in this work are represented by traffic datasets collected from divergent locations at different time. It is assumed that datasets representing different scenarios are non-i.i.d. For this assumption, Section \ref{subsec_ds} will intuitively demonstrate the divergence of selected scenarios. This paper also proposes a method in Section \ref{subsec_div} to quantitatively measure the divergence between distributions of motion data from different scenarios. The following of this sub-section will generally formulate the trajectory prediction task in the interactive scenario.

In an interactive scenario, driving behaviors and motions of vehicles influence each other. The interaction-aware trajectory predictor observes historical trajectories of vehicles in the scene over $t_h$ seconds. Then, it predicts future trajectories of target vehicles over $t_f$ seconds. In detail, the input to the predictor is:
\begin{equation}
{\bf{X }}{\rm{ =  }}\left[ {{\bf{t}}{{\bf{r}}^{(t - {t_h})}},...,{\bf{t}}{{\bf{r}}^{(t - 1)}},{\bf{t}}{{\bf{r}}^{(t)}}} \right].\label{eq_inputs}
\end{equation}
In (\ref{eq_inputs}), ${\bf{t}}{{\bf{r}}^{\left( t \right)}} = \left[ {x_0^{\left( t \right)},y_0^{\left( t \right)},...,x_i^{(t)},y_i^{(t)},...,x_n^{\left( t \right)},y_n^{\left( t \right)}} \right]$ represents $x$ and $y$ co-ordinates of vehicles at time $t$. ${\bf{t}}{{\bf{r}}^{(t)}}$ includes $x_0^{\left( t \right)}$ and $y_0^{\left( t \right)}$, representing co-ordinates of the target vehicle (predicted vehicle). $x_i^{\left( t \right)}$ and $y_i^{\left( t \right)},\left( {i = 1,...,n} \right)$ are co-ordinates of surrounding vehicles. Since decision-making expects the prediction to provide all possible future motions~\cite{zhan2018probabilistic}, the output is not a single predicted trajectory of target vehicle. Instead, the output is considered as the estimated bi-variate distribution over:
\begin{equation}
{\bf{Y}} = \left[ {{\bf{tr}}_0^{\left( {t + 1} \right)},...,{\bf{tr}}_0^{\left( {t + {t_f}} \right)}} \right],
\end{equation}
with ${\bf{tr}}_0^{(t)} = \left[ {x_0^{\left( t \right)},y_0^{\left( t \right)}} \right]$ are future co-ordinates of the target vehicle. Thus, the output can be formulated as:
\begin{equation}
{\mathop{\rm P}\nolimits} ({\bf{Y}}|{\bf{X}})\sim{\bf{\Gamma }}.\label{eq_outputs}
\end{equation}
where ${\bf{\Gamma }} = \left[ {{{\bf{\Gamma }}^{(t + 1)}},...,{{\bf{\Gamma }}^{(t + {t_f})}}} \right]$ are parameters of a bivariate Gaussian distribution at each time step over the prediction horizon. After introducing the fundamental task, the CL task for trajectory prediction in continuous scenarios is described in Section \ref{subsection_cl_traj}.

\subsection{Continual Learning Task for Trajectory Prediction}
\label{subsection_cl_traj}
CL problem consists of a sequence of tasks coming in a stream~\cite{clrobot_lesort2020continual, cossu2021continual_survey2}. In this work, tasks refer to trajectory predictions in interactive scenarios, which is formulated in Section \ref{subsection_traj}. Due to the limited data and computing resource, it is assumed that all data are not available at once. Moreover, the observed data cannot be entirely stored. Under these assumptions, the trajectory predictor is expected to perform well consistently in sequential tasks, i.e., predictions in continuous scenarios. Important concepts of the CL tasks in this work are defined as follow: 
\subsubsection{Continuous Scenarios}
Continuous scenarios consist of a sequence of datasets which can be characterized as $S = \{ {d_1},...,{d_c},...{d_n}\}$. In $S = \{ {d_1},...,{d_c},...{d_n}\}$, datasets are collected at different time and locations representing divergent scenarios. The AV is assumed to pass these scenarios orderly, and ${d_i} \in S,\left( {i = 1,...,n} \right)$ is the $i^{th}$ scenario to be observed by the AV. It should be noted that scenarios are allowed to appear in $S$ more than one times, corresponding to the situation that the AV passes the same scenario again.
\subsubsection{Current Scenario}
The current scenario refers to the newly coming scenario $d_c$ in $S$. The data of the current scenario are fully available to train the trajectory predictor. The trajectory prediction in the current scenario corresponds to the current task in CL.
\subsubsection{Past Scenarios}
In continuous scenarios $S$, the scenarios ${d_i} \in S,\left( {i = 1,..,c - 1} \right)$ that have been observed are termed past scenarios. The data of past scenarios cannot be directly used for training, while the CL strategy can store these data with limited amounts. The trajectory predictions in past scenarios refer to previous tasks in CL.
\subsubsection{Future Scenarios}
The scenarios ${d_j} \in S,\left( {j = c+1,..,n} \right)$ to be arrived are termed future scenarios. Data of future scenario are not available. The next "current scenario" will come from future scenarios. The number of scenarios $n$ can be unknown.

\subsubsection{Catastrophic Forgetting}
Catastrophic forgetting in this work refers to the phenomenon that the predicting accuracy in past scenarios declines after the predictor learns the data of current scenario.

 After learning the data of current scenario, a trajectory predictor is expected to perform well among all scenarios that have been learned. For example, in continuous scenarios $S = \{ {d_1},...,{d_c},...,{d_n}\}$, the data of the current scenario ${d_c} \in S$ are fully accessible for training. Conversely, the data of past scenarios ${d_i} \in S,\left( {i = 1,..,c - 1} \right)$ are not fully accessible. The performance of prediction is evaluated on all testing sets from ${d_i} \in S,\left( {i = 1,..,c} \right)$.
Denoting the predicting error in the $i^{th}$ scenario as $R_i$, the aim of CL task for trajectory prediction in continuous scenarios is to minimize the average errors $R$ of all scenarios that have been learned in $S$:
\begin{equation}
    {\rm{minimize\;}}R = \frac{1}{c}\sum\limits_{i = 1}^c {{R_i}}.
\end{equation}

\section{Dynamic Gradient Scenario Memory For Vehicle Trajectory Prediction}
\label{D-GSM-section}

This section introduces the proposed CL approach for vehicle trajectory prediction in continuous scenarios, termed Dynamic Gradient Scenario Memory (D-GSM). As shown in Fig. \ref{fig_method}, D-GSM consists of four modules: 1) a scenario repository, 2) a traffic divergence measuring module, 3) a dynamic memory module, and 4) a dynamic memory-aware continual learning module.

Since the old data is not directly accessible, the scenario repository is used to store the observed data of past scenarios as memory data. When a new scenario comes, the scenario repository updates by storing new data with abandoning a part of old memory data due to the limited storage. Then, the traffic divergence between the current and past scenarios is measured using the memory data from the scenario repository. Next, according to measuring results, the dynamic memory module dynamically allocates memory data into the CL module. Finally, with the help of allocated memory data, the CL module trains the trajectory predictor using GEM strategy. The details of four modules in D-GSM are presented in Section \ref{subsec_repo}, Section \ref{subsec_div}, Section \ref{subsec_dynamicmem}, and Section \ref{subsec_cl}, respectively.

\subsection{Scenario Repository}
\label{subsec_repo}
The scenario repository is used to store observed data with a limited amount as memory data. In $S = \{ {d_1},...,{d_c},...,{d_n}\}$, when a new scenario $d_c$ arrives, the scenario repository updates by storing the data from the current scenario $d_c$. Since the storage space of the repository is not infinite, it also abandons a part of memory data stored from $d_i (i=1,…,c-1)$. Supposing the upper limit of data storage is $M$, amount of memory data corresponding to the $i^{th}$ scenario will be $m_i=M /c$. Thus, when a new scenario arrives, $m_i$ declines while $c$ increases.

In D-GSM, the memory data stored in the scenario repository are used to measure the traffic divergence between scenarios. Then, according to the measurement, the dynamic memory module allocates memory data into continual learning modules for model training. Different tasks are allocated with different amounts of memory data to improve the training efficiency.

\subsection{Traffic Divergence Measuring}
\label{subsec_div}
The divergence between different scenarios is the main reason causing catastrophic forgetting of trajectory prediction in continuous scenarios. On the contrary, if the new scenario is quite similar with the observed one, the catastrophic forgetting may not happen. Moreover, exploiting similarity of tasks is also a key factor in CL~\cite{kudithipudi2022biological-nature}. Thus, the divergence (or described as similarity) measurement between traffic scenarios is considered in this work. In detail, traffic divergence measuring module in D-GSM uses a KLD-based metric to quantify the traffic divergence between scenarios. The measurement estimates the similarity between CL tasks, enabling the strategy of data allocation in dynamic memory module. 

According to previous studies~\cite{bao2021lifelong, yu2020trafficspatiotemporal}, the divergence between interactive scenarios can be represented by the difference of spatiotemporal dependency among vehicles. From the aspect of scenario data, the spatiotemporal dependency can be formulated as the conditional probability density function (CPDF) $ p \left( {{\bf{Y}}|{\bf{X}}} \right)$, where ${\bf{Y}}$ represents the future trajectories of the predicted vehicle, and ${\bf{X}}$ represents the observed historical trajectories of all vehicles. 

Supposing $p_1$ and $p_2$ are the CPDFs of two scenarios. The distance between CPDFs are calculated by conditional Kullback-Leibler divergence (CKLD):
\begin{equation}
\begin{array}{l}
{\mathop{\rm CKLD}\nolimits} \left( {{p_1}\left( {{\bf{Y}}|{\bf{X}}} \right)||{p_2}\left( {{\bf{Y}}|{\bf{X}}} \right)} \right)\\
 = \int {{p_1}\left( {\bf{X}} \right)\int {\log \left( {\frac{{{p_1}\left( {{\bf{Y}}|{\bf{X}}} \right)}}{{{p_2}\left( {{\bf{Y}}|{\bf{X}}} \right)}}} \right)} {p_1}\left( {{\bf{Y}}|{\bf{X}}} \right)d{\bf{Y}}d{\bf{X}},} 
\end{array}\label{eq_ckld}
\end{equation}

 In (\ref{eq_ckld}), $p_1$ and $p_2$ are approximated by estimations of Gaussian Mixture Models (GMMs) from a Mixture Density Networks (MDN)~\cite{bishop1994MDN}. It should be noted that, in the implementation, to facilitate the learning process, the condition ${\bf{X}}$ with fixed dimensions is used, which only considers the closest $N$ surrounding vehicles. Besides, to represent the interactions of vehicles, instead of directly using historical trajectories, $k$ eigenvectors of a 2D Laplacian matrix concatenated with historical trajectories of the predicted vehicle ${{\bf{X}}_0} = \left[ {{\bf{tr}}_0^{(t - {t_h})},...,{\bf{tr}}_0^{(t - 1)}} \right]$ are used to generate the condition ${\bf{X}}$. To distinguish this processed condition ${\bf{X}}$ from the historical trajectories described in (\ref{eq_inputs}), we denote this condition one as ${\bf{X}}_{cond}^N$:
\begin{equation}
   {\bf{X}}_{cond}^N = [{{\bf{X}}_0},{{\bf{v}}_1},...,{{\bf{v}}_k}],\label{eq_conditionX}
\end{equation}
where ${{\bf{v}}_i}(i = 1,...,k)$ are eigenvectors of a 2D Laplacian matrix. 

Supposing that $e\left( {{\bf{tr}}_i^k,{\bf{tr}}_j^k} \right)$ is Euclidean distance between the $i^{th}$ vehicle and $j^{th}$ vehicle at time $k$, and $\lambda$ is a decay parameter, the Laplacian matrix $\bf{L}_p$ is computed by:
\begin{equation}
    {\bf{L}_p} = {\bf{D}} - {\bf{A}}
\end{equation}
where ${\bf{A}} = {\left( {{a_{i,j}}} \right)_{N \times N}}$ and ${\bf{D}} = {\left( {{d_{i,j}}} \right)_{N \times N}}$. The elements in matrix ${\bf{A}}$ and ${\bf{D}}$ are calculated by:
\begin{equation}
\begin{array}{l}
{a_{i,j}} = \exp \left( { - \sum\limits_{k = t - {t_h}}^{t - 1} {{\omega _k}e\left( {{\bf{tr}}_i^k,{\bf{tr}}_j^k} \right)/\sum\limits_{k = t - {t_h}}^{t - 1} {{\omega _k}} } } \right),\\
{\omega _k} = {\lambda ^{(t - 1) - k}},k = t - {t_h},...,t - 1,\\
{d_{i,j}} = \left\{ \begin{array}{l}
\sum\limits_{j = 1}^N {{a_{i,j}},i = j} \\
0,i \ne j
\end{array} \right.,
\end{array}
\end{equation}
After estimating the GMM for each condition ${\bf{X}}_{cond}^N$, Monte-Carlo sampling is used to compute the Kullback-Leibler divergence (KLD) since the KLD between two GMMs is not analytically attractable. Then CKLD is obtained based on KLD. Specifically, assuming that distribution ${p_1}({\bf{X}}_{cond}^N)$ has $n_1$ samples denoted as ${\bf{X}}_{cond,i}^N(i = 1,...,{n_1})$, $n_{mc}$ samples denoted as ${{\bf{Y}}_j}(j = 1,...,{n_{mc}})$ are sampled from ${p_1}({\bf{Y}}|{\bf{X}}_{cond,i}^N)$, KLD between two distributions is obtained by Monte-Carlo sampling:
\begin{equation}
\begin{array}{l}
{\mathop{\rm KLD}\nolimits} \left( {{p_1}\left( {{\bf{Y}}|{\bf{X}}_{cond,i}^N} \right)||{p_2}\left( {{\bf{Y}}|{\bf{X}}_{cond,i}^N} \right)} \right)\\
 = \frac{1}{{{n_{mc}}}}\sum\limits_{j = 1}^{{n_{mc}}} {\left( {\log {p_1}\left( {{{\bf{Y}}_j}|{\bf{X}}_{cond,i}^N} \right) - \log {p_2}\left( {{{\bf{Y}}_j}|{\bf{X}}_{cond,i}^N} \right)} \right)}.
\end{array}
\end{equation}Finally, CKLD is calculated as:
\begin{equation}
\begin{array}{l}
{\mathop{\rm CKLD}\nolimits} \left( {{p_1}\left( {{\bf{Y}}|{\bf{X}}_{cond}^N} \right)||{p_2}\left( {{\bf{Y}}|{\bf{X}}_{cond}^N} \right)} \right)\\
 = \frac{1}{{{n_1}}}\sum\limits_{i = 1}^{{n_1}} {{\mathop{\rm KLD}\nolimits} \left( {{p_1}\left( {{\bf{Y}}|{\bf{X}}_{cond,i}^N} \right)||{p_2}\left( {{\bf{Y}}|{\bf{X}}_{cond,i}^N} \right)} \right)}.
\end{array}\label{eq_ckld_ip}
\end{equation}

Moreover, since the original KLD is calculated using memory data in the scenario repository with a storage limit, the amount of memory data for each scenario decreases when a new scenario arrives, as stated in Section \ref{subsec_repo}. The influence of data usage on CKLD is also considered in this paper. More investigations and the basic requirement of this calculation are represented in Appendix \ref{appendix-kld}.

CKLD can preliminarily reveal the divergence between scenarios since large CKLD indicates a significant distance between GMMs. If two scenarios are the same, CKLD will equal zero. However, CKLD is asymmetrical, which means that if the $p_1$ and $p_2$ are exchanged in (\ref{eq_ckld}), CKLD will be different. The asymmetry may lead to contradictory measuring results when comparing divergence among traffic scenarios. Therefore, this work proposes a novel metric termed weighted-CKLD to address the problem mentioned above. The weighted-CKLD is calculated as:
\begin{equation}
{\rm{CKLD}}_{1,2}^{wt} = {w_1}{\rm{CKLD}}\left( {{p_1}||{p_2}} \right) + {w_2}{\rm{CKLD}}\left( {{p_2}||{p_1}} \right)
\label{eq_wtckld}
\end{equation}
In (\ref{eq_wtckld}), $w_1$ and $w_2$ are weights ($w_1+w_2=1$). And $p_1$ and $p_2$ in (\ref{eq_wtckld}) represent CPDFs of scenarios to compare. Which weight to be larger depends on which distribution is highlighted. For example, there are three CPDFs $\left \{p_1, p_2, p_3\right \}$ corresponding to three scenarios. We want to compare the divergence between $\left \{p_1, p_2\right \}$ and  $\left \{p_1, p_3\right \}$. The weighted-CKLD $\rm{CKLD}_{1,2}^{wt}$ and $\rm{CKLD}_{1,3}^{wt}$ need to be calculated. In this case, the highlighted one is $p_1$. Then we choose larger value for $w_1$ in $\rm{CKLD}_{1,2}^{wt}$ and $\rm{CKLD}_{1,3}^{wt}$. 

\begin{figure}[tp]
      \centering
      \includegraphics[scale=1.0]{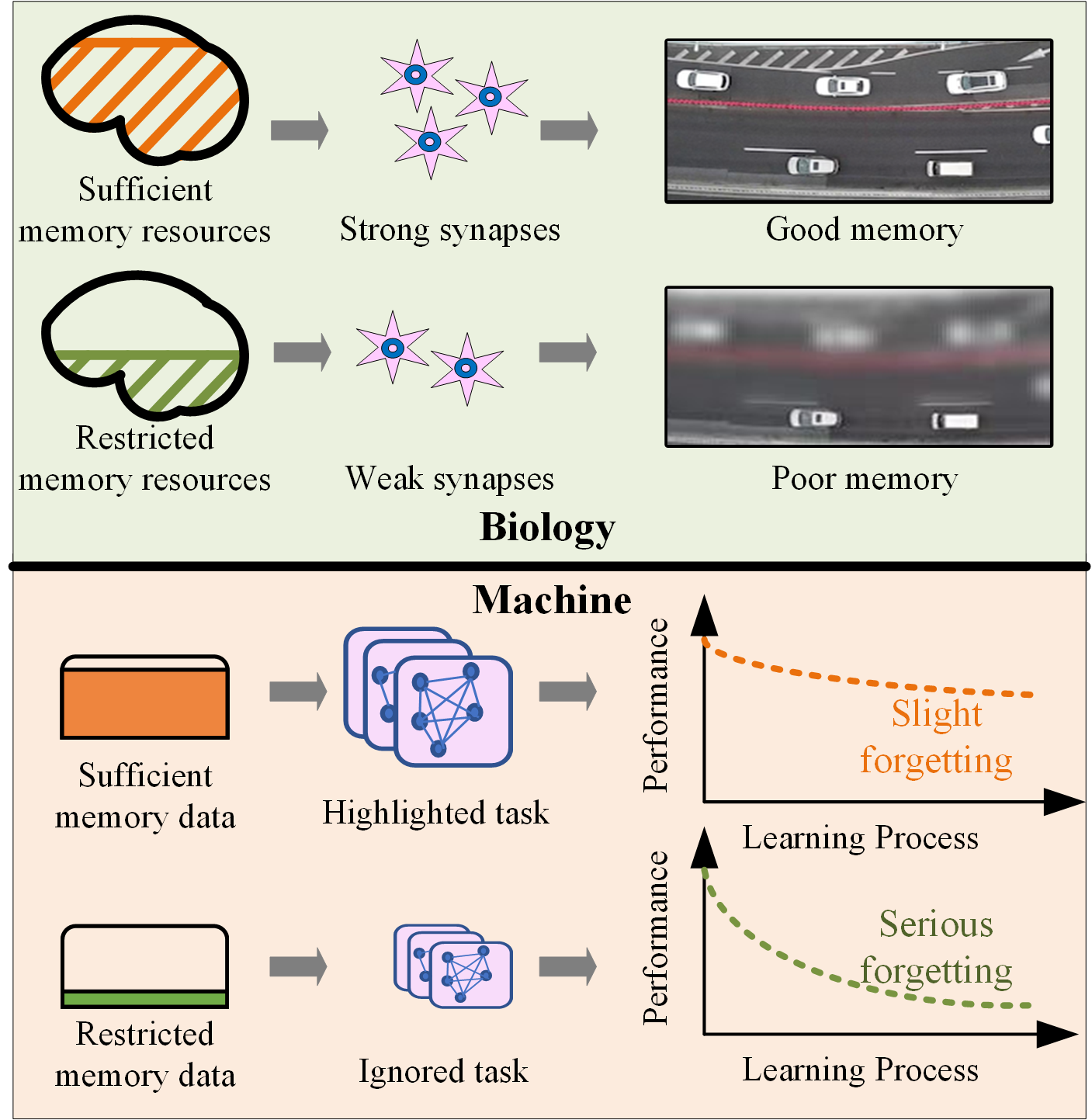}
      \captionsetup{font={small}}
      \caption{Dynamic memory in D-GSM is inspired by synaptic theory in brain science. Sufficient memory resources help agents to remember specific issues/ tasks. Restricted memory resources bring rapid forgetting.}
      \label{fig_idea}
   \end{figure}

In our work, D-GSM explores the relationship between catastrophic forgetting and the spatiotemporal dependency divergence between current traffic scenarios and past scenarios. Thus the highlighted scenario is the current one. Weighted-CKLD measuring the traffic divergence is utilized to improve the training efficiency of CL module by applying the dynamic memory, which is introduced in Section \ref{subsec_dynamicmem}.

\subsection{Dynamic Memory Module}
\label{subsec_dynamicmem}

The dynamic memory module allocating different memory data for divergent tasks to improve the efficiency of CL is inspired by the synaptic theory of memory in brain science~\cite{langille2018synaptic}. The synaptic theory indicates that the forgetting process can become rapid when memory resources are restricted~\cite{finnie2012metaplasticity, kudithipudi2022biological-nature}. The idea inspired by brain science and the allocating principle in the dynamic memory module are shown in Fig. \ref{fig_idea}. In this work, the memory resources refer to the memory data. Using more memory data may enable the model to have better retention. However, more memory data also burden the training cost in CL. 

Based on the inspiration, an allocation strategy for memory data is developed in the dynamic memory module to make a trade-off between the performance and training cost: Intuitively, compared with tasks in pretty different scenarios, tasks in past scenarios that are similar to the current one are "easier" for models to remember since the learning-based model updates parameters depending on the training data. Therefore, to balance the effectiveness and efficiency of CL, the dynamic memory in D-GSM allocates more memory data for observed tasks that are "hard" to remember. 

In detail, results of traffic divergence measuring is used to enable the dynamic memory module to dynamically allocate memory data from the scenario repository to the CL module. Instead of allocating memory data to all previous tasks equally as GEM~\cite{gem}, the dynamic memory of D-GSM allocates a different number of memory data to different previous tasks. The specific amounts of allocated data to a past scenario depend on its traffic divergence to the current scenario. Firstly, the traffic divergence measuring module uses memory data stored in the scenario repository to calculate weighted-CKLD. Then the maximum weighted-CKLD is selected:
\begin{equation}
\rm{CKLD}_{\max }^{wt} = \max \left\{ {\rm{CKLD}_{c,r}^{wt}|r \in \left\{ {1,...,c - 1} \right\}} \right\}\label{eq_max}
\end{equation}
where $\rm{CKLD}_{c,r}^{wt}$ is the weighted-CKLD between the current scenario and the $r^{th}$ past scenario. The traffic scenario corresponding to the maximum weighted-CKLD indicates the largest divergence with the current scenario. Denote the maximum amounts of memory data for CL as $M_{cl} (M_{cl}\le M)$. Then, memory data with the most amounts ${m_{\max }} = \frac{{{M_{cl}}}}{{c - 1}}$ is allocated to this traffic scenario. Finally, memory data amounts allocated to the $r^{th}$ scenario is calculated by:
\begin{equation}
{m_r^{\rm{D}}} = {m_{\max}}\frac{{\rm{CKLD}_{c,r}^{wt}}}{{\rm{CKLD}_{\max }^{wt}}}.\label{eq_dmem}
\end{equation}
Equation (\ref{eq_dmem}) presents the specific data allocation of dynamic memory, which will be used to formulate the dynamic memory-aware CL module of D-GSM in Section \ref{subsec_cl}.

\subsection{Dynamic Memory-Aware Continual Learning}
\label{subsec_cl}
The allocated memory data from the dynamic memory module are used to apply the CL strategy in model training. Inspired by GEM~\cite{gem}, the CL module in D-GSM firstly defines loss functions for previous tasks, i.e., trajectory predictions in past scenarios. Then, inequality constraints are set to interface the training process, where the model observes the training data of the current scenario. Finally, with the help of the Quadratic Program (QP) algorithm, proposed gradients satisfying the inequality constraints are applied to update parameters, which avoids the increment of losses on previous tasks.

The loss functions for previous tasks are calculated using memory data. Original GEM treats all previous tasks equally without considering the similarity between tasks, which may bring an unnecessary computational burden. As introduced in Section \ref{subsec_dynamicmem}, instead of treating all tasks equally, D-GSM uses dynamic memory data allocation to construct loss functions for previous tasks:
\begin{equation}
l\left( {{f_{\bm{\theta }}},{m_r^{\rm{D}}}} \right) = \frac{1}{{{m_r^{\rm{D}}}}}\sum\limits_{i = 1}^{{m_r^{\rm{D}}}} {l\left( {{f_{\bm{\theta }}}\left( {{{\bf{X}}_i},r} \right),{{\bf{Y}}_i}} \right)}\label{eq_pastloss}
\end{equation}
where $m_r^{\rm{D}} (r=1,..,c-1)$ is the allocated amounts of samples described in (\ref{eq_dmem}). ${f_{\bm{\theta }}}$ is the vehicle trajectory predicting model parameterized by ${\bm{\theta }}$, and $\left( {{{\bf{X}}_i},r,{{\bf{Y}}_i}} \right)$ is the $i^{th}$ sample in the allocated memory data corresponding to the ${r^{th}} \left( {r = 1,..,c - 1} \right)$ past scenario. Then, in the training process of the current task, losses in (\ref{eq_pastloss}) are used 
to define inequality constraints to avoid the increment of losses on previous tasks. 

Supposing that $\left( {{\bf{X}},c,{\bf{Y}}} \right)$ are samples of the current task, and $f_{\bm{\theta }}^{'}$ represents the predicting model state at the end of learning of the last traffic scenario, the inequality constraints are formulated as:
\begin{equation}
\begin{array}{l}
{\rm{minimiz}}{{\rm{e}}_{\bm{\theta }}}{\rm{\; }}l\left( {{f_{\bm{\theta }}}\left( {{\bf{X}},c} \right),{\bf{Y}}} \right)\\
{\rm{s}}{\rm{.t}}{\rm{. \quad}}l\left( {{f_{\bm{\theta }}},{m_r^{\rm{D}}}} \right) \le l\left( {f_{\bm{\theta }}^{'},{m_r^{\rm{D}}}} \right){\rm{ ,for\; all\; }} r< c.
\end{array}\label{eq_constraint}
\end{equation}

For an efficient implementation, this paper assumes that the loss function is local linear (when learning rate is small), and denotes loss gradients of the current and previous tasks as ${\bf{g}}$ and ${{\bf{g}}_r}$, respectively. (\ref{eq_constraint}) can be rephrased into:
\begin{equation}
\begin{array}{l}
\left\langle {{\bf{g}},{{\bf{g}}_r}} \right\rangle : = \left\langle {\frac{{\partial l\left( {{f_{\bm{\theta }}}\left( {{\bf{X}},c} \right),{\bf{Y}}} \right)}}{{\partial {\bm{\theta }}}},\frac{{\partial l\left( {{f_{\bm{\theta }}},{m_r^{\rm{D}}}} \right)}}{{\partial {\bm{\theta }}}}} \right\rangle  \ge 0{\rm{, }}\\
{\rm{for\;all\; }}r < c.
\end{array}\label{eq_contraints_grad}
\end{equation}If constraints (\ref{eq_contraints_grad}) are satisfied, the proposed gradient $\bf{g}$ to update parameters will not increase the loss of previous tasks. Otherwise, the gradient $\bf{g}$ will be projected to the closest gradient ${\bf{\tilde g}}$ (in squared L2 norm) satisfying all constraints in (\ref{eq_contraints_grad}):
\begin{equation}
    \begin{array}{l}
{\rm{minimiz}}{{\rm{e}}_{{\bf{\tilde g}}}}\quad \frac{1}{2}\left\| {{\bf{g}} - {\bf{\tilde g}}} \right\|_2^2\\
s.t.{\rm{ \quad}}\left\langle {{\bf{\tilde g}},{{\bf{g}}_r}} \right\rangle  \ge 0,{\rm{ for\;all \;}}r < c.
\end{array}\label{eq_projection}
\end{equation}To solve (\ref{eq_projection}) efficiently, a QP algorithm is used. The primal of a QP with inequality constraints is described as:
\begin{equation}
\begin{array}{l}
{\rm{minimiz}}{{\rm{e}}_z}{\rm{\quad  }}\frac{1}{2}{{\bf{z}}^{\rm{T}}}{\bf{Cz}} + {{\bf{p}}^{\rm{T}}}{\bf{z}}\\
{\rm{s}}{\rm{.t}}{\rm{.\quad}}{\bf{Qz}} \ge {\bf{b}}
\end{array}\label{eq_primal}
\end{equation}
where $\mathbf{C} \in \ \mathbb{R}  ^{p\times p} ,\mathbf{p} \in \mathbb{R} ^{p} ,\mathbf{Q} \in \mathbb{R} ^{(c-1)\times p}$ and $\mathbf{b} \in \mathbb{R} ^{c-1}$. The dual problem of (\ref{eq_primal}) is:
\begin{equation}
\begin{array}{l}
{\rm{minimiz}}{{\rm{e}}_{u,v}}{\rm{\quad}}\frac{1}{2}{{\bf{u}}^{\rm{T}}}{\bf{Cu}} - {{\bf{b}}^{\rm{T}}}{\bf{v}}\\
{\rm{s}}{\rm{.t}}{\rm{.\quad}}{{\bf{Q}}^{\rm{T}}}{\bf{v}} - {\bf{Cu}} = {\bf{p}},\\
{\rm{                     }}{\bf{v}} \ge 0.
\end{array}\label{eq_dual}
\end{equation}

\begin{figure}[tp]
      \centering
      \includegraphics[scale=1.0]{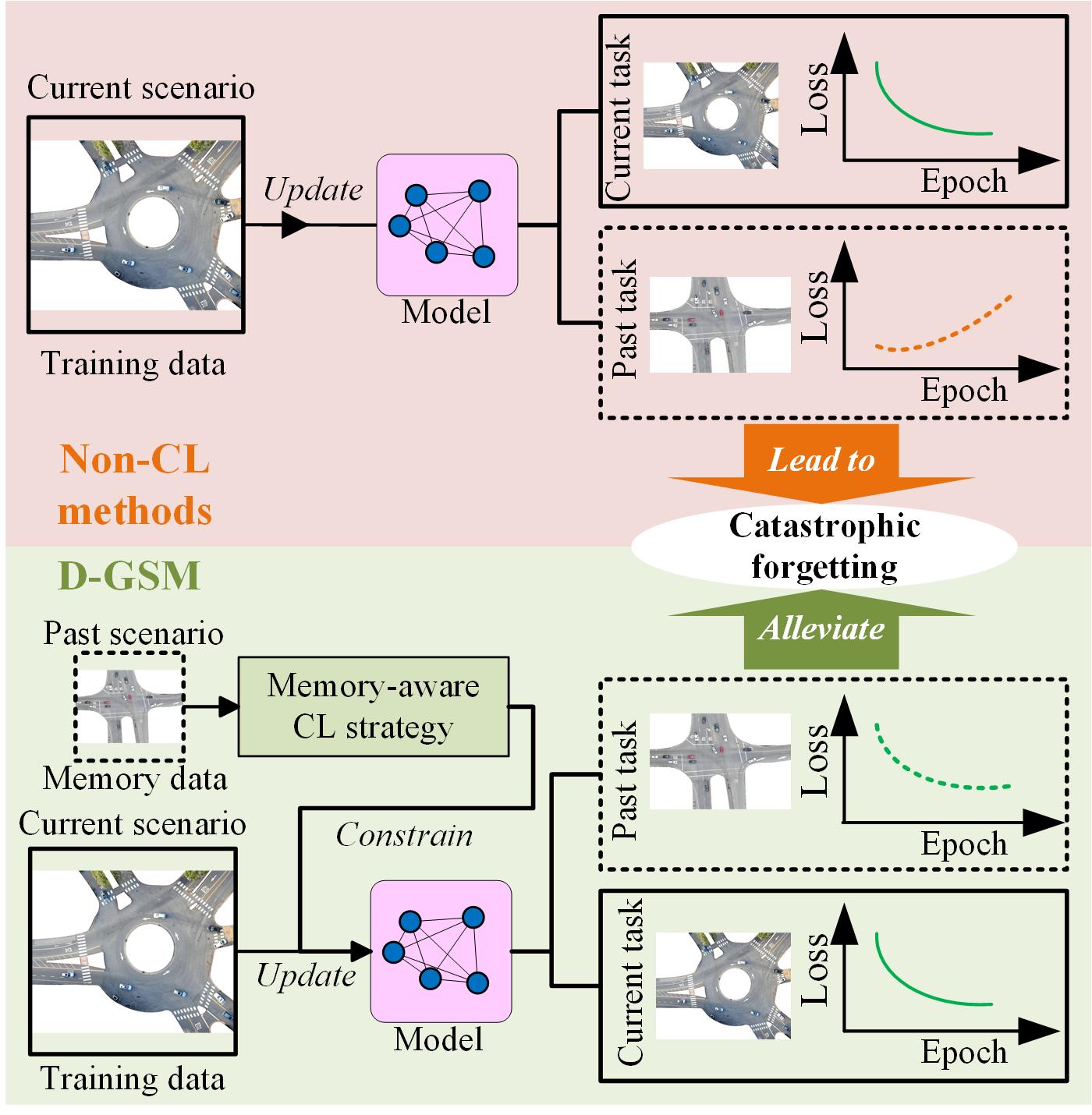}
      \captionsetup{font={small}}
      \caption{The mechanism of D-GSM to realize continual learning in continuous scenarios. Compared with non-CL methods, dynamic memory-aware continual learning strategy of D-GSM constrains the increment of loss on previous tasks, which alleviates catastrophic forgetting.}
      \label{fig_mech}
\end{figure}

If $\left( {{{\bf{u}}^ * },{{\bf{v}}^ * }} \right)$ is a solution to (\ref{eq_dual}), then there will be a solution ${{\bf{z}}^ * }$ satisfying ${\bf{C}}{{\bf{z}}^ * } = {\bf{C}}{{\bf{u}}^ * }$. Thus, the primal GEM QP to (\ref{eq_contraints_grad}) can be described as:
\begin{equation}
   \begin{array}{l}
{\rm{minimiz}}{{\rm{e}}_{\bf{z}}}{\rm{\quad}}\frac{1}{2}{{\bf{z}}^T}{\bf{z}} - {{\bf{g}}^T}{\bf{z}} + \frac{1}{2}{{\bf{g}}^T}{\bf{g}}\\
{\rm{s}}{\rm{.t}}{\rm{.\quad}}{\bf{Gz}} \ge 0,
\end{array}\label{eq_gem_primal}
\end{equation}
where ${\bf{G}} =  - \left[ {{{\bf{g}}_1},...,{{\bf{g}}_{c - 1}}} \right]$ and the constant term ${{\bf{g}}^T}{\bf{g}}$ is discarded. (\ref{eq_gem_primal}) is a QP on p variables (the number of trainable parameters of trajectory predictor). The dual of it is formulated as:
\begin{equation}
   \begin{array}{l}
{\rm{minimiz}}{{\rm{e}}_{\bf{v}}}{\rm{\quad}}\frac{1}{2}{{\bf{v}}^{\rm{T}}}{\bf{G}}{{\bf{G}}^{\rm{T}}}{\bf{v}} + {{\bf{g}}^{\rm{T}}}{{\bf{G}}^{\rm{T}}}{\bf{v}}{\rm{  }}\\
{\rm{s}}{\rm{.t}}{\rm{.\quad}}{\bf{Gz}} \ge 0.
\end{array}\label{eq_gem_dual}
\end{equation}
Once the dual problem (\ref{eq_gem_dual}) is solved for ${{\bf{v}}^ * }$, the projected gradient update can be recovered as ${\bf{\tilde g}} = {{\bf{G}}^{\rm{T}}}{{\bf{v}}^ * } + {\bf{g}}$. According to~\cite{gem}, adding a small constant $\gamma  \ge 0$ to ${{\bf{v}}^ * }$ biased the gradient projection to updates that favored beneficial backwards transfer in practice. Fig. \ref{fig_method} shows the complete process of D-GSM. Fig. \ref{fig_mech} also shows the mechanism of D-GSM to realize continual learning. Compared with not-CL methods, D-GSM applies the memory-aware continual learning strategy to avoid the loss on previous tasks during the current updating.

The entire algorithm of D-GSM is summarized in Algorithm \ref{al_1}. To clearly explain the input and output in each step, symbols in  including $f_{repo}$ for repository updating, $f_{div}$ for divergence measuring, $f_{alloc}$ for dynamic memory module, $f_{cl}$ for CL training strategy and $l_{prev}$, $l_{cur}$ for previous and current losses are newly used in Algorithm \ref{al_1}.

\begin{algorithm}{}
\renewcommand{\algorithmicrequire}{\textbf{Inputs:}}
\renewcommand{\algorithmicensure}{\textbf{Outputs:}}
\caption{Dynamic Gradient Scenario Memory (D-GSM) for Vehicle Trajectory Prediction in Continuous Scenarios}\label{al_1}
\begin{algorithmic}[1]
\REQUIRE Data of continuous scenarios $S$ = \{$d_1$,..., $d_c$, ..., $d_n$\}; memory limit $M$, $M_{cl}$; functions $f_{repo}$, $f_{div}$, $f_{alloc}$, $f_{cl}$; loss $l_{prev}$, $l_{cur}$; model ${f_{\bm{\theta }}}$.
\ENSURE Prediction of the $r^{th}$ scenario ${\mathop{\rm P}_{r}\nolimits} ({\bf{Y}}|{\bf{X}})\sim{\bf{\Gamma }}$ in (\ref{eq_outputs}), for $r$=1:c.
\FOR{current scenario $c=1$ to $n$}
\FOR{$i=1$ to $c$}
\STATE Memory data $d_i^{'}$ $\gets$ $f_{repo}(d_i, m_i)$. $\triangleright$ Scenario repository updating. Amount $m_i=M/c$.
\IF{$c>1$}
\FOR{$r=1$ to $c-1$}
\STATE $\rm{CKLD_{c,r}^{wt}}$ $\gets$ $f_{div}(d_c^{'},d_r^{'})$ $\triangleright$ Traffic divergence measuring in (\ref{eq_conditionX})-(\ref{eq_wtckld}). 
\STATE Calculate $\rm{CKLD}_{\max }^{wt}$ as (\ref{eq_max}). 
\STATE $m_r^{\rm{D}}$ $\gets$ $f_{alloc}(\rm{CKLD}_{\max }^{wt}, M_{cl}, \rm{CKLD_{c,r}^{wt}})$. $\triangleright$ Dynamic memory allocating as (\ref{eq_dmem}).
\STATE $l_{prev}$ $\gets$ $l({f_{\bm{\theta }}}, m_r^{\rm{D}})$ $\triangleright$ Previous loss in (\ref{eq_pastloss}).
\STATE ${\bm{\theta }}$ $\gets$ $f_{cl}(l_{prev}, l_{cur})$ $\triangleright$ CL strategy in (\ref{eq_constraint})-(\ref{eq_gem_dual}).
\ENDFOR
\ELSE
\STATE Model training in a single scenario (c==1).
\ENDIF
\STATE Evaluate ${f_{\bm{\theta }}}$ with the testing set of $d_i$.
\ENDFOR
\ENDFOR
\end{algorithmic}
\end{algorithm}

\section{Experiments}
\label{sec_experiments}
\begin{figure}[tp]
      \centering
      \includegraphics[scale=1.0]{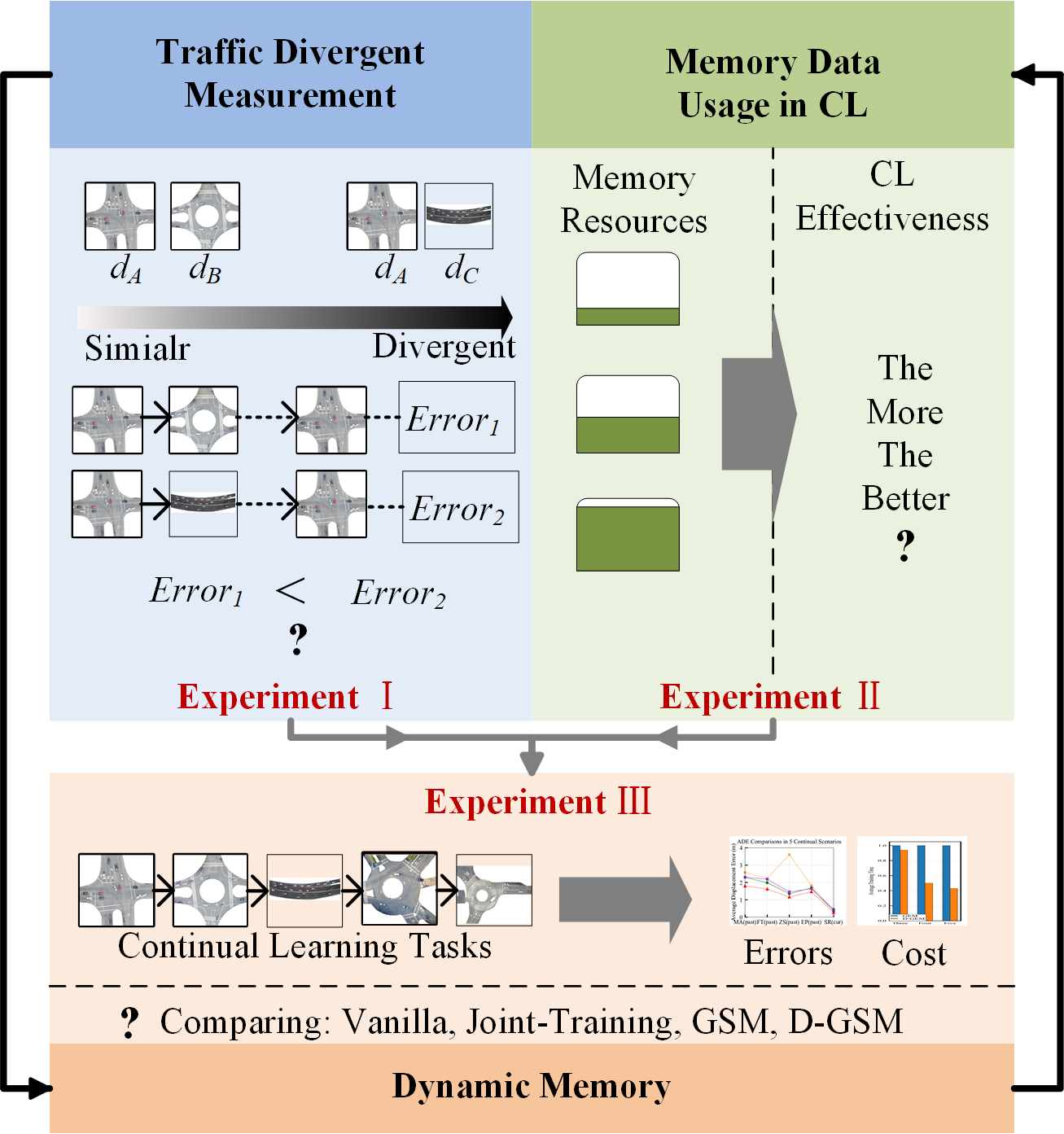}
      \captionsetup{font={small}}
      \caption{The logic of three experiments in this paper. The experiment \uppercase\expandafter{\romannumeral1} explores the relationship between catastrophic forgetting and traffic divergence. After exploring the reasons for forgetting, the experiment \uppercase\expandafter{\romannumeral2} investigates the memory capability of the proposed CL approach using different amounts of memory data. The proposed D-GSM utilizes measurements of traffic divergence to dynamically allocate memory data. Experiment \uppercase\expandafter{\romannumeral3} compares the performance of D-GSM with several baselines.}
      \label{fig_logic}
\end{figure}

To investigate CL problems formulated in Section \ref{section_formulation} and evaluate the proposed approach introduced in Section \ref{D-GSM-section}, three experiments using datasets representing different scenarios are conducted. As shown in Fig. \ref{fig_logic}, the first experiment explores the relationship of catastrophic forgetting and task similarity of trajectory prediction in continuous scenarios. The second experiment investigates the influence of memory resource on the CL performance. The comparison between non-CL models and models applied with the proposed CL approach is demonstrated in the third experiment, evaluating the performance of D-GSM in continuous scenarios. This section firstly introduces used datasets and experimental settings. Then, experimental results with analysis are presented.

\begin{table}[bp]
    \centering
    \captionsetup{font={small}}
    \caption{Different Scenarios from INTERACTION Dataset Used In Experiments}
    \begin{tabular}{c c c}
    \toprule
 Scenario Name & Interactive Scenario Type & Notation\\ \midrule
 DR$\_$USA$\_$Intersection$\_$MA & Urban Intersection & $d_1$ \\
 DR$\_$USA$\_$Roundabout$\_$FT & Urban Roundabout & $d_2$ \\
 DR$\_$CHN$\_$Merging$\_$ZS & Ramp Merging & $d_3$ \\
 DR$\_$USA$\_$Roundabout$\_$EP & Urban Roundabout &  $d_4$ \\
 DR$\_$USA$\_$Roundabout$\_$SR & Urban Roundabout & $d_5$ \\
    \bottomrule
    \end{tabular}
    \label{table_datasets}
\end{table}

\begin{figure*}[htbp]
      \centering
      \includegraphics[scale=0.8]{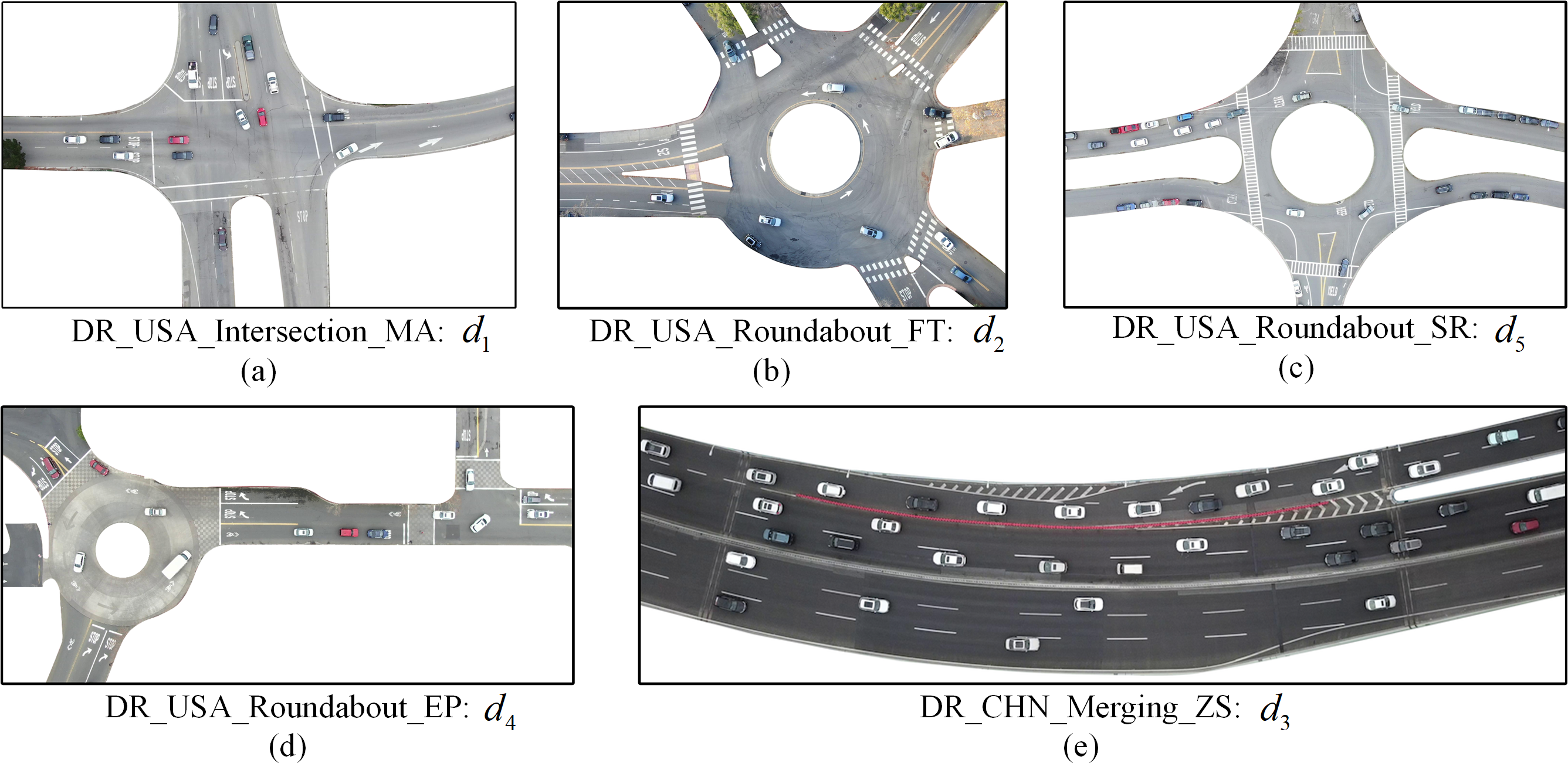}
      \captionsetup{font={small}}
      \caption{Intuitive visual aids for scenarios: Various interactive driving scenarios are used in experiments based on INTERACTION datasets. The first two letters (”DR”) of names represent that the data are collected by drones. The following three letters represent the corresponding country (”USA” is for the U.S.A., and ”CHN” is for China). The last two letters are scenario codes to distinguish different locations~\cite{zhan2019interaction-dataset}.}
      \label{fig_maps}
\end{figure*}

\subsection{Datasets and Implementations}
\label{subsec_ds}
Since this study focuses on CL tasks for interactive behavior learning, typical interactive scenarios, including merging scenarios in highways, intersections, and roundabout scenarios in urban areas, are selected as examples to evaluate the proposed approach. Specifically, experiments are conducted based on INTERACTION~\cite{zhan2019interaction-dataset}, which collect data from different locations at different time to represent divergent scenarios. For convenience, these scenarios are denoted with different notations, as shown in Table \ref{table_datasets}.

\begin{table}[bp]
    \centering
    \captionsetup{font={small}}
    \caption{Implemented Key Parameters}
    \setlength{\tabcolsep}{3.5mm}{
    \begin{tabular}{c c c c}
    \toprule
 Parameters & & & Implemented Values \\ \midrule
 Observation horizon & & & 2s \\
 Prediction horizon & & & 4s\\ 
 Input Size & & & 2 \\
 Output Size & & & 5 \\
 Number of ST-GCNN Layers & & & 1 \\
 Number of TXP-CNN Layers & & & 5 \\
 Number of Training Epochs & & & 250 \\
 Learning Rate & & & 0.001 \\
 Kernel Size & & & 3 \\
    \bottomrule
    \end{tabular}}
    \label{table_parameters}
\end{table}

Fig.\ref{fig_maps} also demonstrates the photos of these scenarios intuitively. In detail, as shown in Fig.~\ref{fig_maps}(a), $d_1$ is a busy all-way-stop intersection with lanes controlled by stop signs. $d_2$ is demonstrated by Fig.\ref{fig_maps} (b), which is a busy 7-way roundabout with one ”yield” branch and six ”stop” branches. As Fig.\ref{fig_maps} (e) shows, $d_3$ is collected on a highway that contains several sub-scenarios. The upper two lanes are zipper merging, and it is a ramp for the middle two lanes. Moreover, it is a forced merging for the lower three lanes where vehicles must change lanes. Fig.\ref{fig_maps} (c) and (d) show the $d_5$ and $d_4$, which are roundabouts controlled by stop signs and yield signs, respectively.

\begin{figure}[htbp]
      \centering
      \includegraphics[scale=1.0]{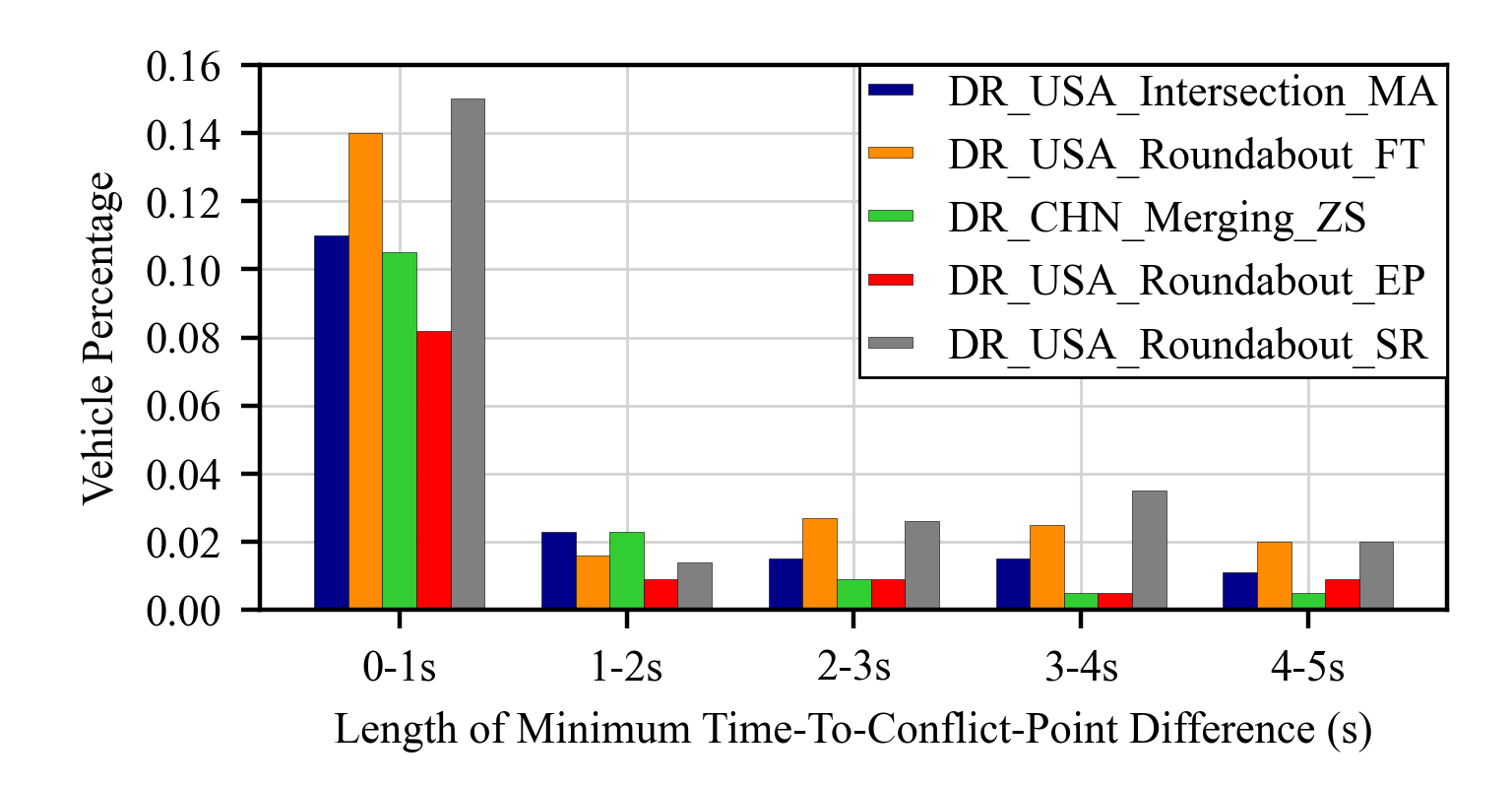}
      \captionsetup{font={small}}
      \caption{Comparisons for distributions of the minimum time-to-conflict-point among selected scenarios. The x-axis represents the length of minimum time-to-conflict-point ($\triangle TTCP_{min}$) in seconds. And the y-axis are the percentage of vehicles that with particular $\triangle TTCP_{min}$ relative to total number of vehicles in the dataset. The higher values of y-axis represent larger density of interactive behaviors ~\cite{zhan2019interaction-dataset}. Detailed definition of minimum time-to-conflict is described in Appendix \ref{appendix-ttcp}.}
      \label{fig_tccp}
\end{figure}

These scenarios are divergent from many aspects such as road geometry, traffic participants, traffic rules, driving behaviors of vehicles, etc. Fig.\ref{fig_tccp} also demonstrates the comparison of the distribution of the minimum time-to-conflict-point($\triangle TTCP_{min}$). $\triangle TTCP_{min}$ represents the density of interactive behaviors among these scenarios~\cite{zhan2019interaction-dataset}. As shown in Fig. \ref{fig_tccp}, it can be found that selected five scenarios have different density of interactive behaviors. More details on the definition of $\triangle TTCP_{min}$ are provided in Appendix \ref{appendix-ttcp}. To quantify the divergence of scenarios, as described in \ref{subsec_div}, this paper mainly focuses on the different spatiotemporal dependencies among vehicles between scenarios. And the weighted-CKLD is proposed to measure the divergence of distribution in the aspect of spatiotemporal dependencies between these non-i.i.d datasets which represent divergent scenarios.

In data processing, extracted features include IDs and coordinates of vehicles with timestamps. Then, trajectories samples are processed for vehicle trajectory prediction and traffic divergence measuring, respectively, as described in Section \ref{subsection_traj} and Section \ref{subsec_div}. In detail, the closest 5 surrounding vehicles are considered ($N=5$) in the conditional ${\bf{X}}_{cond}^N$ for measurements of the traffic divergence. Furthermore, the number of eigenvector $k$ in (\ref{eq_conditionX}) is set as 3. The training strategy of continual trajectory prediction is that the learned model is loaded to be trained on new datasets without directly accessing old data from past scenarios when a new scenario comes. Approximate 10,000 data samples are used from each scenario. The selected dataset corresponding to each scenario is split into training, validation, and testing datasets by 7:1:2.

\begin{figure*}[htbp]
      \centering
      \includegraphics[scale=1.0]{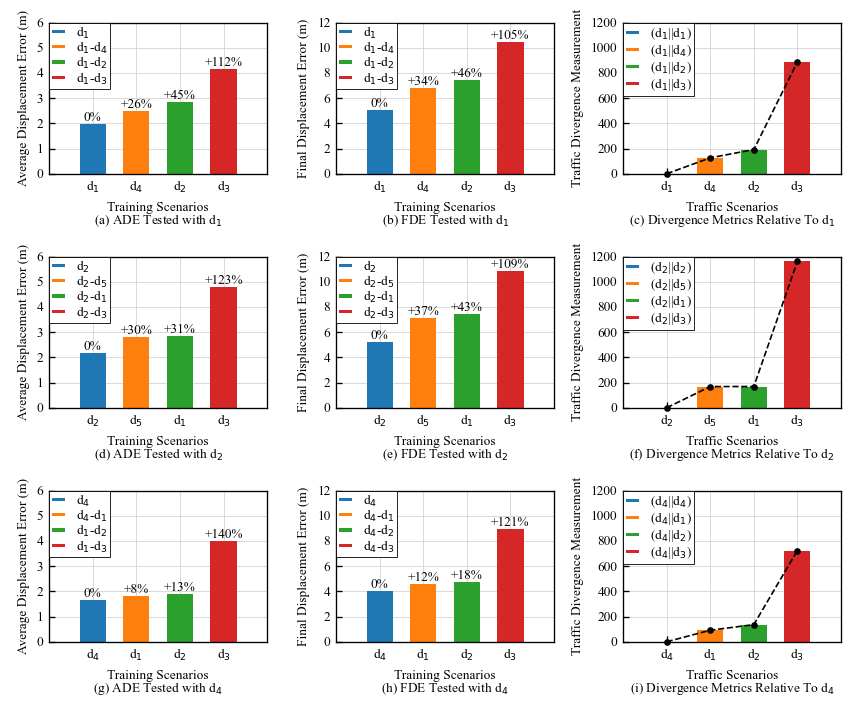}
      \captionsetup{font={small}}
      \caption{Catastrophic forgetting compared with the measurement of traffic divergence.}
      \label{fig_wtckld}
\end{figure*}

The proposed approach is appropriate for interaction-aware models using gradient descent-based methods for updating parameters. In this work, the Social-STGCNN model~\cite{s-stgcnn_mohamed2020social}, a graph-based interaction-aware trajectory predictor, is adopted as the base model. In Social-STGCNN, interactions between vehicles are represented by a graph where nodes represent the positions of vehicles, and edges with a weighted matrix encode the spatial interdependency between vehicles. Then, graph representations are inputted into the spatiotemporal graph convolutional neural networks (ST-GCNN). Based on the operation of ST-GCNN, a time-extrapolator convolutional neural network (TXP-CNN) outputs the predicted distribution of trajectories as (\ref{eq_outputs}).

The model observes 2s historical trajectories and predicts trajectories for 4s. They are trained for 250 epochs using Stochastic Gradient Descent (SGD), and the learning rate is set as 0.001. Detailed implementation of models are shown in Table \ref{table_parameters}. The training loss function is the negative log-likelihood:
\begin{equation}
   {\mathop{l}\nolimits} \left( {\bm{\theta }} \right) =  - \sum {\log \left( {{\rm P}\left( {{\bf{Y}}|{\bf{\hat \Gamma }}} \right)} \right)}
\end{equation}
where ${\bf{\hat \Gamma }}$ is the estimated Gaussian distribution over future trajectories ${\bf{Y}}$. Models are implemented using PyTorch\footnote{https://pytorch.org}.

The metrics for predicting accuracy are Average Displacement Errors (ADE) and Final Displacement Errors (FDE), commonly used in trajectory predictions~\cite{alahi2016social-lstm, s-stgcnn_mohamed2020social, li2021hierarchicalgraph}. Suppose each test set has ${N_{ts}}$ samples, the $i^{th}$ sample of predicted co-ordinates of target vehicle and the ground truth at time $t$ are denoted as ${\bf{\hat tr}}_{0,i}^{(t)}$ and ${\bf{tr}}_{0,i}^{(t)}$. ADE represents the average Euclidean distance for the whole predicted trajectory:
\begin{equation}
ADE = \frac{{\sum\nolimits_{i = 1}^{{N_{ts}}} {\sum\nolimits_{t' = t + 1}^{t + {t_f}} {{{\left\| {{\bf{\hat tr}}_{0,i}^{(t')} - {\bf{tr}}_{0,i}^{(t')}} \right\|}_2}} } }}{{{N_{ts}} \times {t_f}}}.
\end{equation}
FDE represents the Euclidean distance for the final predicted position:
\begin{equation}
   FDE = \frac{{\sum\nolimits_{i = 1}^{{N_{ts}}} {{{\left\| {{\bf{\hat tr}}_{0,i}^{(t + {t_f})} - {\bf{tr}}_{0,i}^{(t + {t_f})}} \right\|}_2}} }}{{{N_{ts}}}}.
\end{equation}

\begin{figure*}[bp]
      \centering
      \includegraphics[scale=1.0]{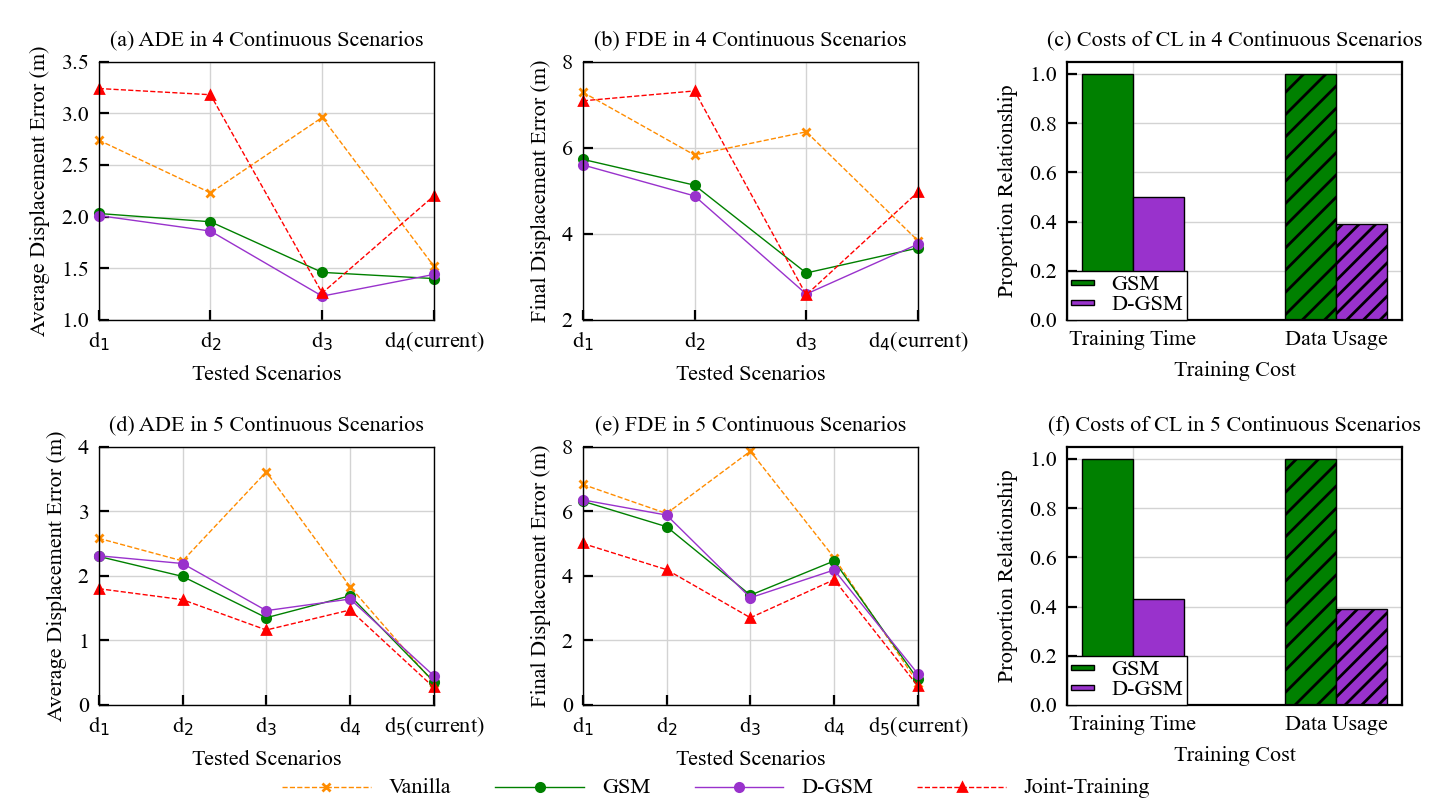}
      \captionsetup{font={small}}
      \caption{Trajectory predictions in continuous scenarios: The orange cross markers with dash lines represent the predicting performance vanilla social-STGCNN model (base model). The red triangle markers with dash lines represent the performance of base model with joint-training, which has fully access to all traffic data. And the solid lines are proposed continual learning method, where the green one represents the GSM (ours) and the purple one represents the D-GSM (ours).}
      \label{fig_traj}
\end{figure*}
   
\subsection{Experimental Settings}
This work mainly investigates the catastrophic forgetting of vehicle trajectory prediction in continuous scenarios. Instead of treating all previous tasks equally as GEM, the proposed D-GSM uses dynamic memory for CL. Dynamic memory allocates more memory resources, i.e., the memory data for tasks that are easier to be forgotten to balance the effectiveness and efficiency. D-GSM judges different tasks by measuring traffic divergence between the current task and the previous ones. To validate the rationality of dynamic memory and evaluate the performance of D-GSM, three experiments are conducted based on INTERACTION dataset. Experiments \uppercase\expandafter{\romannumeral1} and \uppercase\expandafter{\romannumeral2} investigate key factors of predictions in continuous scenarios and also validate the rationality of our design of dynamic memory. Experiment \uppercase\expandafter{\romannumeral3} evaluates the proposed approach. Fig. \ref{fig_logic} demonstrates the logic of experiments in this work.

In detail, experiment \uppercase\expandafter{\romannumeral1} explores the relationship between catastrophic forgetting and traffic divergence. Base models are firstly trained with continuous scenarios consisting of two scenarios. After learning the second scenario, the model is tested with the firstly learned one. Compared to the test after learning the first scenario, the increment of predicting errors reflects the catastrophic forgetting. Then, the traffic divergence between different scenarios is presented by the weighted-CKLD. Finally, the relationship between catastrophic forgetting and traffic divergence is analyzed.

After investigating influencing factors of forgetting in experiment \uppercase\expandafter{\romannumeral1}, experiment \uppercase\expandafter{\romannumeral2} explores the retention of the proposed approach from the aspect of memory resources. The performance of the proposed approach with different memory data is compared. Models with the proposed CL approach are trained and tested in continuous scenarios. To make a clear comparison between different memory usage, models allocate 100, 500, and 1,000 samples of memory data for each previous task in the CL strategy without using the dynamic memory. The average performance over continuous scenarios is compared.

Experiment \uppercase\expandafter{\romannumeral3} aims at evaluating the performance of the proposed approach. Since the dynamic memory allocating is expected to be compared to the model with equal memory allocating, three groups of continuous scenarios are set, including ${S_{three}} = \{ {d_1},{d_2},{d_3}\}$, ${S_{four}} = \{ {d_1},{d_2},{d_3},{d_4}\}$, and ${S_{five}} = \{ {d_1},{d_2},{d_3},{d_4},{d_5}\}$. Each group of continuous scenarios has more than two scenarios (In continuous scenarios with only two scenarios, the memory allocation of D-GSM is the same as GSM.) The model settings are:
\begin{itemize}
    \item \textbf{Vanilla}: The base model (Social-STGCNN) without applying continual learning approach.
    \item \textbf{GSM (ours)}: The base model applied with the proposed approach but without dynamic memory. The memory usage for all past scenarios are equally set as ${m_{\max }} = {M_{cl}}/\left( {c - 1} \right)$. In this experiment, ${M_{cl}}$ are 3,500 samples.
    \item \textbf{D-GSM (ours)}: The base model applied with the entire proposed approach. The memory usage for the $r^{th}$ past scenario is described in (\ref{eq_dmem}), $M_{cl}$ are also set as 3,500 samples.
    \item \textbf{Joint-Training}: Vanilla base model with joint training. It is a non-continual learning setting that does not follow the storage limitation assumptions of continual learning. Data from all scenarios are available at once. Models are trained with mixed data from all scenarios.
\end{itemize}

   \begin{table}[tp]
    \centering
    \captionsetup{font={small}}
    \caption{Average Predicting ADE(m)/ FDE(m) in Continuous Scenarios with Different Number of Memory Data}
    \begin{tabular}{c c c c c}
    \toprule
 \makecell{Continuous\\Scenarios} & \makecell{0\\Per Task} & \makecell{100\\Per Task}  & \makecell{500\\Per Task} & \makecell{1,000\\Per Task} \\ \midrule
    $\{d_1,d_2\}$ & 2.33/5.89 & 2.29/5.65 & 2.12/5.51 & \textbf{1.99/5.14} \\ 
    $\{d_1, d_2,d_3\}$ & 3.61/8.43 & 2.30/5.46 & 2.01/4.84 & \textbf{1.87/4.58}  \\ 
    $\{d_1,d_2,d_3,d_4\}$ & 2.36/5.83 & 1.91/4.75 & 1.78/4.54&\textbf{1.70/4.33}  \\ 
    \bottomrule
    \end{tabular}
    \label{table_mem}
\end{table}
   
  \begin{table*}[tp]
    \centering
    \captionsetup{font={small}}
    \caption{Vehicle Trajectory Predicting Performance (ADE / FDE) In Continuous Scenarios}
    \begin{tabular}{c|c|c c c c}
    \toprule
 Continuous Scenarios & Tested Scenario & \makecell{Vanilla} & \makecell{Joint Training} & \makecell{GSM (ours)} &\makecell{ D-GSM (ours)} \\ \midrule
    \makecell{Continuous Scenarios $S_{three}$:\\{$d_1,d_2, d_3$}}  &\makecell{The $1^{st}$ past scenario: $d_1$\\The $2^{nd}$ past scenario: $d_2$\\Current scenario: $d_3$\\ (Average)}  &  \makecell{4.82 / 11.53 \\ 5.29 / 12.15 \\ \textbf{0.73} / \textbf{1.62} \\(3.61 / 8.43)} & \makecell{3.15 / 7.80 \\ 2.69 / \textbf{6.43}  \\ 1.23 / 2.54 \\ (2.36 / 5.59)}  &\makecell{\textbf{2.55} / 6.36 \\ \textbf{2.96} / 6.84 \\ \textbf{0.80} / \textbf{1.74} \\ (\textbf{2.10} / \textbf{4.98})}  & \makecell{2.67 / \textbf{6.14} \\ 3.05 / 6.91 \\ 0.90 / 1.97 \\ (2.21 / 5.01)}   \\ \midrule
    
    \makecell{Continuous Scenarios $S_{four}$:\\{$d_1,d_2, d_3, d_4$}}  &\makecell{The $1^{st}$ past scenario: $d_1$\\ The $2^{nd}$ past scenario: $d_2$\\The $3^{rd}$ past scenario: $d_3$\\ Current scenario: $d_4$ \\ (Average)}  &  \makecell{2.74 / 7.29 \\ 2.23 / 5.83 \\ 2.96 / 6.37 \\ 1.52 / 3.84 \\ (2.36 / 5.83)} & \makecell{3.24 / 7.09 \\ 3.18 / 7.32 \\ 1.26 / \textbf{2.58} \\ 2.20 / 4.97 \\ (2.47 / 5.49)} &\makecell{2.03 / 5.73 \\ 1.95 / 5.13 \\ 1.46 / 3.09 \\ \textbf{1.40} / \textbf{3.67} \\(1.71 / \textbf{4.04})}  & \makecell{\textbf{2.01} / \textbf{5.60} \\ \textbf{1.86} / \textbf{4.88} \\ \textbf{1.23} / 2.60 \\ 1.44 / 3.76 \\(\textbf{1.63} / 4.21)}   \\ \midrule
    
    \makecell{Continuous Scenarios $S_{five}$:\\{$d_1,d_2, d_3, d_4, d_5$}}  &\makecell{The $1^{st}$ past scenario: $d_1$\\ The $2^{nd}$ past scenario: $d_2$\\The $3^{rd}$ past scenario: $d_3$\\ The $4^{th}$ past scenario: $d_4$ \\ Current scenario: $d_5$ \\(Average)}  &  \makecell{2.58 / 6.83 \\ 2.23 / 5.93 \\ 3.61 / 7.86 \\ 1.83 / 4.56 \\ \ 0.34 / 0.67 \\ (2.12 / 5.17)} & \makecell{ \textbf{1.80} / \textbf{5.00} \\ \textbf{1.63} / \textbf{4.19} \\ \textbf{1.16} / \textbf{2.70} \\ \textbf{1.47} / \textbf{3.88} \\ \textbf{0.27} / \textbf{0.58} \\ (\textbf{1.27} / \textbf{3.27})}   &\makecell{2.30 / 6.31 \\ 1.99 / 5.52 \\ 1.35 / 3.40 \\ 1.69 / 4.46 \\ 0.36 / 0.79 \\(1.54 / 3.05)}  & \makecell{2.31 / 6.35\\ 2.19 / 5.88 \\ 1.46 / 3.32 \\ 1.64 / 4.18 \\ 0.45 / 0.96 \\(1.61 / 4.13)} \\  
    \bottomrule
    \end{tabular}
    \label{table_pred}
\end{table*}

\subsection{Experimental Results and Analysis}
\subsubsection{Experiment \uppercase\expandafter{\romannumeral1}}
In the first experiment, models are tested on the past scenarios that have been observed. Experimental results of the first experiment are shown in Fig. \ref{fig_wtckld}. Taking the first row in Fig. \ref{fig_wtckld} for example, (a) and (b) are ADE and FDE tested on scenario $d_1$ after observing the last scenario in continuous scenarios $S_1=\{{d_1}, {d_4}\}$, $S_2=\{{d_1}, {d_2}\}$, and $S_3=\{{d_1}, {d_3}\}$. The blue bars in (a) and (b) are baselines showing the testing performance of the model trained on $d_1$. (c) shows the weighted-CKLD between new scenarios and the first learned scenario $d_1$. It can be found that, compared to baselines, ADE and FDE increase in all settings after learning the new scenario. The increased predicting errors reveal the catastrophic forgetting of trajectory predictions in continuous scenarios. The highest increment of ADE and FDE occurs in the setting $S_3=\{{d_1}, {d_3}\}$. After learning scenario $d_3$, ADE and FDE increase by 112$\%$ and 105$\%$. Compared to measurements of the traffic divergence represented  by weighted-CKLD in (c), it can be found that the largest weighted-CKLD also corresponds to the setting $S_3 =\{{d_1}, {d_3}\}$. 

The results are similar in the second rows (d)-(f) and third rows (g)-(i). These results are explainable intuitively. First, $d_3$ is a merging type scenario from a highway in China, while $d_1$, $d_2$, and $d_4$ are scenarios belonging to urban areas in the USA. Thus, $d_3$ can be regarded as a divergent scenario to others. As expected, the highest weighted-CKLDs are results between $d_3$ and other scenarios in all groups of Experiment \uppercase\expandafter{\romannumeral1}. Moreover, compared with errors and weighted-CKLD, experimental results indicate that larger traffic divergence brings higher error increments. 
\begin{table}[bp]
    \centering
    \captionsetup{font={small}}
    \caption{Calculated CKLDs Between Different Scenarios}
    \begin{tabular}{c|c c c c c}
    \toprule
    \diagbox{$d'$}{$CKLD(d||d')$}{d} & $d_1$ & $d_2$ & $d_3$ & $d_4$ & $d_5$ \\ \midrule
    $d_1$ & 0 & 151.21 & 87.79 & 65.45 & 121.64 \\
    $d_2$ & 209.30 & 0 & 88.88 & 121.10 & 132.71 \\
    $d_3$ & 1230.01 & 1622.34 & 0 & 987.29 & 1110.29 \\
    $d_4$ & 152.44 & 171.08 & 109.58 & 0 & 97.99\\
    $d_5$ & 269.04 & 183.51 & 93.31 & 164.99 & 0 \\
    \bottomrule
    \end{tabular}
    \label{table_ckld}
\end{table}
\subsubsection{Experiment \uppercase\expandafter{\romannumeral2}}
The second experiment compares the proposed CL approach using different memory data for continual learning. Table \ref{table_mem} shows the average ADE and FDE in three groups of continuous scenarios. The baseline is the vanilla base model which does not use memory data (0 memory data per task) and has the highest average errors (both ADE and FDE) in three tested groups. The model using the most memory data in this experiment has the lowest errors among the three groups of continuous scenarios. It can also be found that in each group, with the increment of memory data allocated to calculate the losses of past scenarios, the average errors decline. These results indicate that using more memory data for losses (\ref{eq_pastloss}) may improve the predicting performance in continuous scenarios. However, using more memory data brings higher training costs. The third experiment will discuss more on the cost, where the training cost of two proposed model settings is compared.

\subsubsection{Experiment \uppercase\expandafter{\romannumeral3}}
The third experiment evaluates the performance of the proposed approach. First, models are continually trained in continuous scenarios. Then models are tested with all scenarios that have learned. This experiment includes three groups of continuous scenarios.Table \ref{table_pred} shows detailed results. Compared to vanilla base model, the proposed GSM and D-GSM models have lower ADE and FDE among three groups of continuous scenarios. Besides, joint-training settings are regarded as the best possible performance in continual learning~\cite{ma2021continual, bao2021lifelong} since all training data are accessible at once in joint training. Joint-training model has the best performance in the continuous scenarios ${S_{five}} = \{ {d_1},{d_2},{d_3},{d_4},{d_5}\}$. However, we also found that in another two groups of continuous scenarios, joint-training models are not the best. The proposed models outperform the joint-training in ${S_{three}} = \{ {d_1},{d_2},{d_3}\}$ and ${S_{four}} = \{ {d_1},{d_2},{d_3},{d_4}\}$. Here, take results in ${S_{three}}$ as example to briefly discuss this "unexpected" phenomenon: 
It may owe to the memory mechanism of the proposed approach and the selected scenarios. In ${S_{three}}$, ${d_3}$ is a highway merging scenario collected in China, while other scenarios are from urban environments in the USA. ${d_3}$ has a larger divergence relative to others. The total number of scenarios is 3, which means that the data of ${d_3}$ has one-third of the training data for joint-training. Besides, the joint-training just mixed all data together to train the model. Due to the divergence, data from ${d_1}$ and ${d_2}$ may "weaken" the performance on ${d_3}$. Meanwhile, data from ${d_3}$ interrupt the performance on ${d_1}$ and ${d_2}$. However, since the proposed approach purely uses memory data to constrain the loss increment on past scenarios, it performs well both in the current and past scenarios.

When the number of scenarios increases as in ${S_{five}}$, on the one hand, highly divergent scenario ${d_3}$ has a lower weight in the whole continuous scenarios. On the other hand, decreasing memory data may reduce the advantage of the proposed methods. As a result, in $S_{five}$, joint-training has the best performance. In summary, these interesting results indicate that the advantage and limitations of the proposed approach are relative to the memory room and the scenario data. There can be some ablative studies to explore more characteristics of the proposed approach in future works.

The ADE and FDE are also detailed in Table \ref{table_pred}.
Besides, CKLDs utilized to calculate weighted-CKLDs for D-GSM are shown in Table \ref{table_ckld}. These experimental results show that the proposed approach improves the performance of vehicle trajectory prediction in continuous scenarios. The catastrophic forgetting of base models is alleviated.

Moreover, the proposed D-GSM aims to improve training efficiency by allocating different memory data to diverse previous tasks. In Fig. \ref{fig_traj}, the green dot markers with solid lines represent the proposed approach without dynamic memory (GSM). The purple dot markers with solid lines represent the proposed D-GSM. Yellow cross markers and red triangle markers with dash lines represent vanilla base model without continual learning strategy and base model with joint-training. Comparisons of training time and the data usage are also demonstrated in Fig. \ref{fig_traj}. Since training time depends on the scale of data and number of epochs, the last column of Fig. \ref{fig_traj} shows the proportional relationships of training time and data usage for two settings. The green bars in Fig. \ref{fig_traj} represent the cost of GSM, and the purple bars represent the D-GSM. (c) show the comparisons in continuous scenarios $S_{four}$, and (f) shows the comparisons in $S_{five}$. The average training time per epoch of GSM is set as 1, which corresponds to approximate 900s. Similarly, for a clear comparison, usage of memory data for GSM is also set as 1, corresponding to 3,500 samples. It is easy to understand that the ratio of time cost is close to the ratio of memory data usage since more data bring more computations. In three groups of continuous scenarios, the time cost and data cost of D-GSM are lower than GSM. The costs are approximate 0.4 to 0.5 times in ${S_{four}}$ and $S_{five}$. It can be found that D-GSM models have a lower time cost without reducing much accuracy.



\section{Conclusions}
This paper proposes a novel continual learning approach termed D-GSM for interactive behavior learning of AVs. D-GSM constrains the loss increment on old tasks when learning a new task. With the help of memory data, the updating strategy for trainable parameters considers both data from the current and previous scenarios at each optimizing step. Therefore, compared to the non-CL approaches, which only focus on learning the current task, the proposed approach mitigates the catastrophic forgetting of interactive behavior learning in continuous scenarios without re-training. Moreover, a novel metric named weighted-CKLD is proposed to measure the traffic divergence between interactive scenarios. By utilizing the measurement of traffic divergent, a dynamic memory for the proposed CL approach is developed to improve the training efficiency. Based on various scenarios datasets from INTERACTION dataset~\cite{zhan2019interaction-dataset}, experiments are conducted to evaluate the proposed approach. Experimental results show that D-GSM outperforms non-CL models in continuous scenarios. Meanwhile, the developed dynamic memory improves the training efficiency by using less memory data. 

The proposed approach strengthens the practicability of AVs, which is beneficial to improve the road safety and traffic efficiency of ITS. Besides, investigations on the uncertainty in traffic or explorations for different factors of traffic divergence may be the potential research area of continual interactive behavior learning for ITS.




%

\small
\bibliographystyle{IEEEtran} 
\bibliography{ref_dgsm}{}

\appendices
\section{Definition of Minimum Time-To-Conflict-Point}
\label{appendix-ttcp}
Referring to \cite{zhan2019interaction-dataset}, the density of interactive behaviors of a scenario can be represented by minimum time-to-conflict-point ($\triangle TTCP_{min}$). $\triangle TTCP_{min}$ is a metric to describe the relative states between two moving vehicles in a scenario, where the paths of the two vehicles share a conflict point but without any forced stop. More details about the concept of the conflict points can be found in \cite{zhan2019interaction-dataset}. Assuming that $TTCP_{i}^{t}=\triangle l_{i}^{t}/v_{i}^{t} (i=1,2)$ is the traveling time to the conflict point of each vehicle in the interactive pairs\cite{zhan2019constructing}. $v_{i}^{t}$ and $\triangle l_{i}^{t}$ are, respectively, the speed of the $i^{th}$ vehicle and its distance to the conflict point along the path at time $t$. Then, $\triangle TTCP_{min}$ is defined as:
\setcounter{equation}{0}
\renewcommand\theequation{A.\arabic{equation}} 
\begin{equation}
    \bigtriangleup TTCP_{min}=\underset{t\in \left [ T_{start},T_{end} \right ] }{min}\left ( TTCP_{1}^{t}-TTCP_{2}^{t} \right )  
\end{equation}
where $T_{start}$ and $T_{end}$ are, respectively, the starting time index when both vehicles appear and the crossing time index when one of the vehicles passes the conflict point. If $\triangle TTCP_{min} \le $3s, then it is defined that interaction exists \cite{zhan2019interaction-dataset, zhan2019constructing}.

\section{Basic Requirement for Calculations of Kullback-Leibler Divergence}
\label{appendix-kld}
 
\begin{figure}[b]
      \centering
      \includegraphics[scale=0.5]{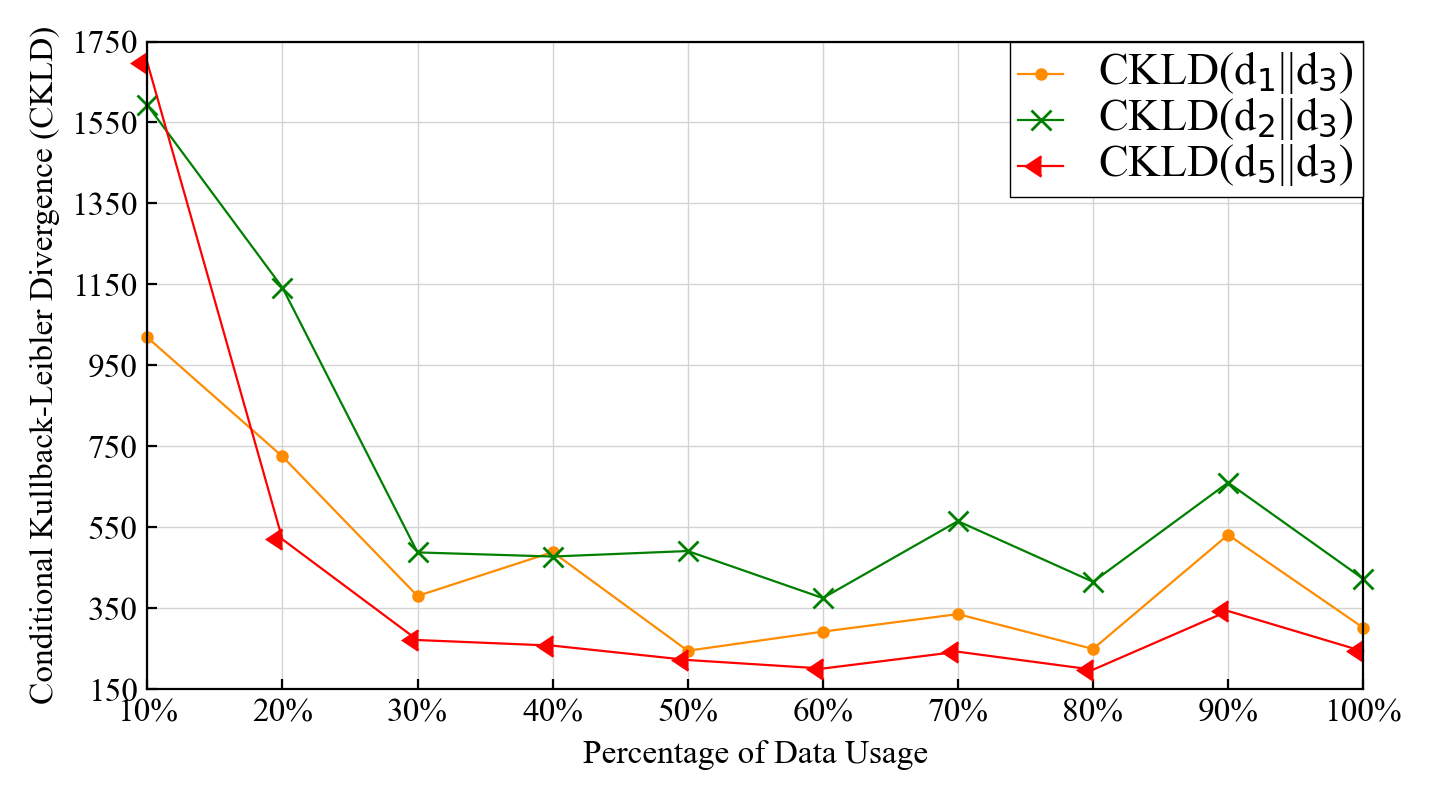}
      \captionsetup{font={small}}
      \caption{The conditional Kullback-Leibler divergence (CKLD) between scenario $d_{3}$ and $d_{i}(i=1,2,5)$ are calculated using different data amounts. The maximum number of processed cases for this experiment is 60,000, which is denoted as 100$\%$ data usage.}
      \label{fig_ckld2data}
\end{figure}

As introduced in Section \ref{subsec_div}, difference of spatiotemporal dependency among vehicles are used to represent divergence between interactive scenarios. In the implementation, the spatiotemporal dependency is formulated as the conditional probability density function (CPDF) over historical and future trajectories of vehicles, and CKLD presented in (\ref{eq_ckld}) is used to measure the distance between two CPDFs. Referring to (\ref{eq_ckld}), $p_{1}$ and $p_{2}$ are assumed as GMMs, which are firstly estimated by MDN~\cite{bishop1994MDN}. Then, Monte-Carlo sampling is applied to calculate the original KLD between estimated GMMs. Finally, CKLD is calculated based on KLD, as formulated in (\ref{eq_ckld_ip}).

The calculation described above takes memory data stored in the scenario repository as inputs. According to Section \ref{subsec_repo}, memory storage space is not infinite. The memory data for each observed scenario decreases when a new scenario comes. However, insufficient data may lead to errors during the calculation of CKLD. In detail, since GMMs are composed of several Gaussian distributions, this unexpected issue may happen when data are insufficient to fit the specific number of Gaussian distributions by MDN. Thus, an experiment is conducted to demonstrate the influence of data usage on the calculation of CKLD. The basic requirement of this calculation will also be provided based on the analysis of experimental results.
 
 \subsection{Experiment Settings}
 \label{app-b-set}
 In the experiment, CKLDs between scenario $d_3$ and the other three scenarios are calculated using different amounts of data. All model settings and assumptions are the same as other experiments conducted in this paper, where GMM to be estimated is composed of 20 Gaussian distributions, and the closest 5 surrounding vehicles with 3 eigenvectors are considered in the conditional ${\bf{X}}_{cond}^N$.

In each group, all scenarios have the same amount of data for the calculation of CKLDs. Processed cases are 60,000 in total. It should be noted that data process refers to the condition ${\bf{X}}_{cond}^N$ and ${\bf{Y}}$ described in Section \ref{subsec_div}, which differs from the processing of trajectories for prediction in this work. Besides, the dimensions of condition ${\bf{X}}_{cond}^N$ for CKLD calculation are different from the processed trajectories to be predicted. The exact amount of raw data can be processed into different amounts of cases for CKLD and trajectory to predict, respectively. Thus, we use "cases" to distinguish "samples," which are used in the paper to represent the processed trajectories for prediction. CKLDs are calculated using different amounts of data as shown in Fig. \ref{fig_ckld2data}. The x-axis is the percentage of data usage in 10 different groups, where $m\%(m=10,20,30,...,100)$ corresponds $m\%$ of total cases(eg: 50$\%$ corresponds 30,000 cases for each scenario).
 
\subsection{Experimental Results and Analysis}
 As shown in Fig. \ref{fig_ckld2data}, results are relatively close most of the time (when the percentage of data usage $\ge 30\%$). The green line representing the CKLD between $d_2$ and $d_3$ is the highest compared with the yellow and red lines. However, when the amount of used data decreases to $20\%$ and $10\%$, CKLD values become much higher than other groups. It should be noted that CKLD between scenario $d_5$ and $d_3$ is the largest when using $10\%$ data, which is converse to other groups. These experimental results indicate that insufficient data (such as the group of $10\%$) may lead to an unreliable calculation since the amount of data is too small to represent the scenario reasonably. Furthermore, we also found that when the percentage of data usage decreases to $5\%$ (or less), CKLD cannot be obtained. This result shows that insufficient data can lead to failed calculations, where MDN cannot estimate the GMMs using inadequate data.
 
 Based on the experimental results, the basic calculation requirement for CKLD under the experimental settings is that the number of processed cases $\ge$ 6,000. This requirement is related to the GMM assumption (how many Gaussian distributions are assumed to be mixed), and the way of data processing (how many surrounding vehicles and eigenvectors are considered in the condition ${\bf{X}}_{cond}^N$). Due to the limitation that insufficient data may not accurately represent the features of spatiotemporal dependencies in a scenario, we also suggest setting a reasonable memory storage space considering the specific learning task. Besides, one of the future works to improve the divergence measuring method is overcoming this limitation.
\vspace{-10 mm}
\begin{IEEEbiography}[{\includegraphics[width=1in,height=1.25in,clip,keepaspectratio]{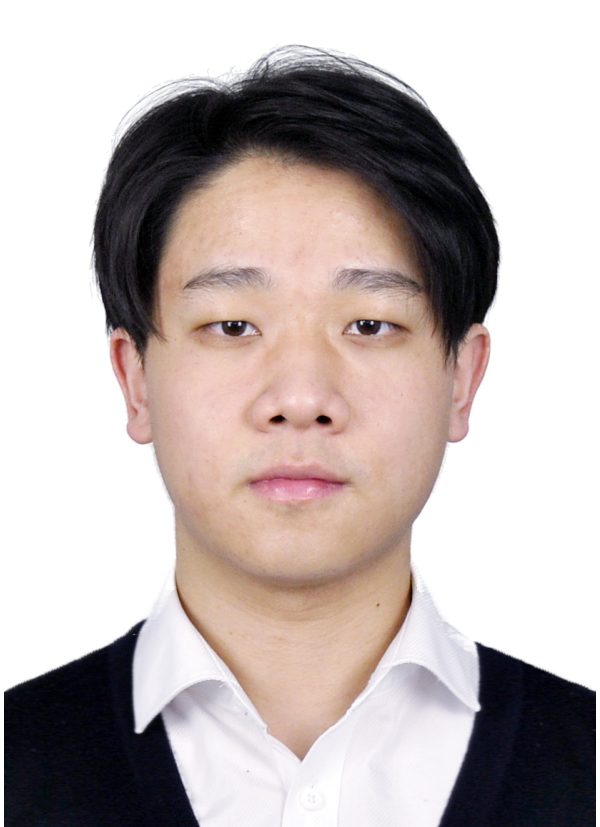}}]{Yunlong Lin}
received the B.S. degree in mechanical engineering from Beijing Institute of Technology (BIT), Beijing, China, in 2022. He is currently pursuing the Ph.D. degree in Beijing Institute of Technology, Beijing, China. His research focuses on interactive behavior modeling, continual learning, and decision-making of intelligent vehicles.
\end{IEEEbiography}
\vspace{-10 mm}
\begin{IEEEbiography}[{\includegraphics[width=1in,height=1.25in,clip,keepaspectratio]{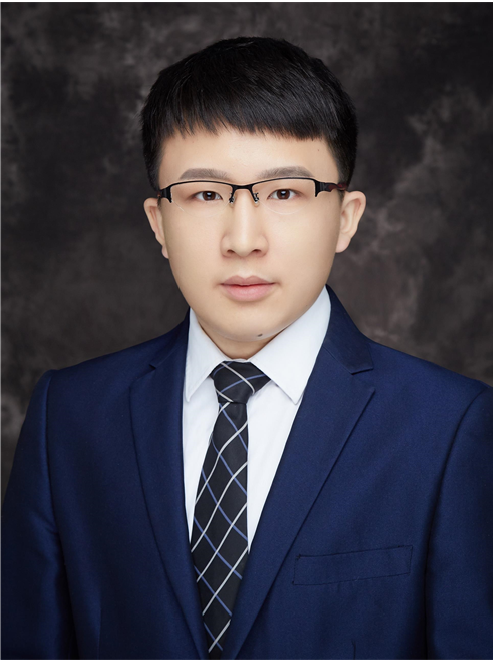}}]{Zirui Li}
received the B.S. degree from the Beijing Institute of Technology (BIT), Beijing, China, in 2019, where he is currently pursuing the Ph.D. degree in mechanical engineering. From June, 2021 to July, 2022, he was a visiting researcher in Delft University of Technology (TU Delft). From Aug, 2022. He is the visiting researcher in the Chair of Traffic Process Automation at the Faculty of Transportation and Traffic Sciences “Friedrich List” of the TU Dresden. His research focuses on interactive behavior modeling, risk assessment and motion planning of automated vehicles.
\end{IEEEbiography}
\vspace{-10 mm}
\begin{IEEEbiography}[{\includegraphics[width=1in,height=1.25in,clip,keepaspectratio]{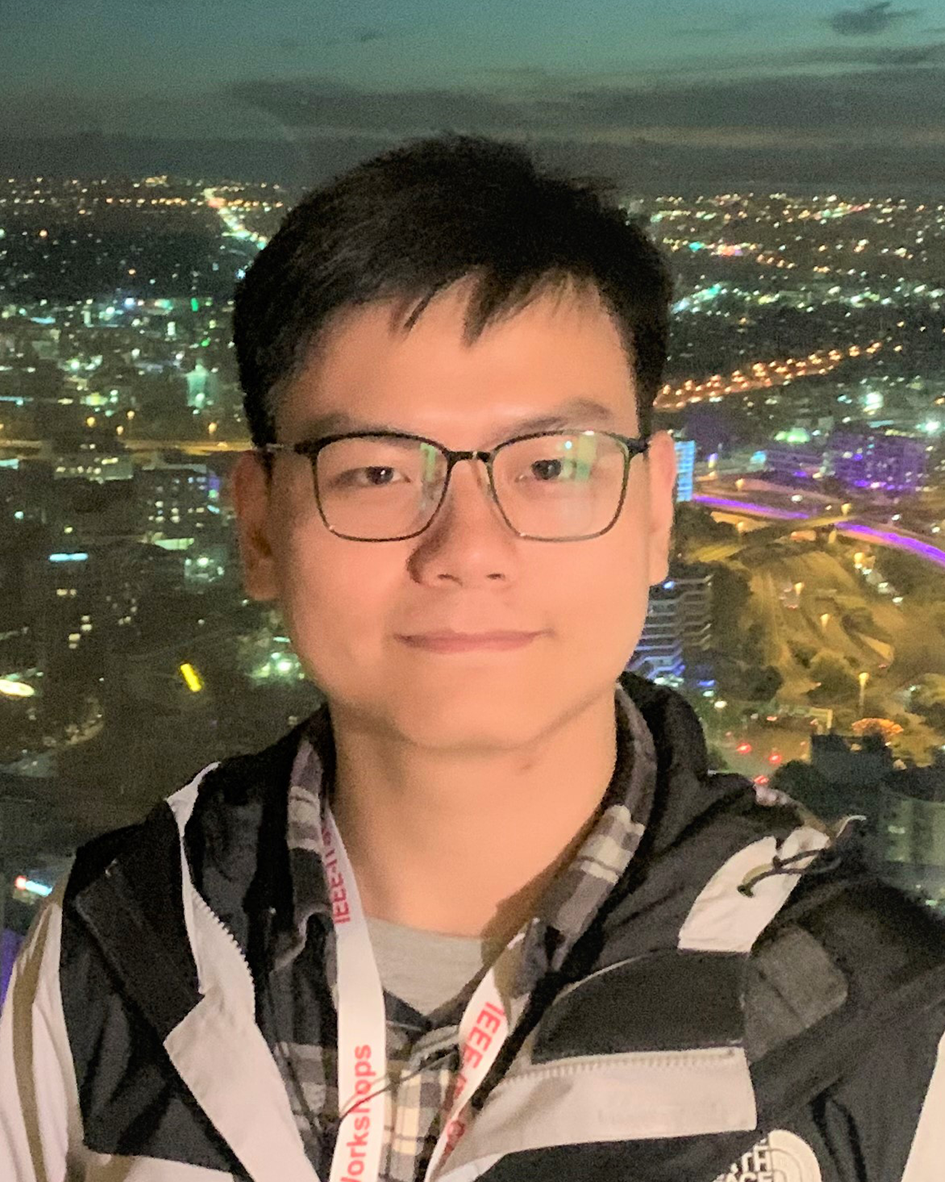}}]{Cheng Gong}
received the B.S. degree in mechanical engineering from Beijing Institute of Technology, China, in 2020. He is currently pursuing the Ph.D. degree in Beijing Institute of Technology, China. His research interests include intelligent vehicles, motion planning and control, machine learning, and life-long learning.
\end{IEEEbiography}
\vspace{-10 mm}
\begin{IEEEbiography}[{\includegraphics[width=1in,height=1.25in,clip,keepaspectratio]{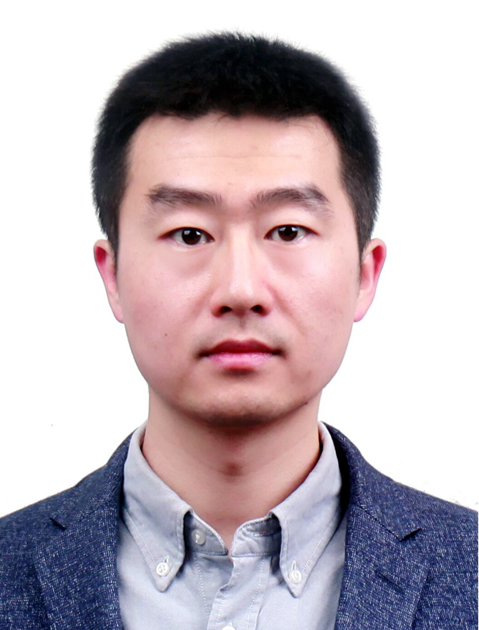}}]{Chao Lu}
received the B.S. degree in transport engineering from the Beijing Institute of Technology (BIT), Beijing, China, in 2009 and the Ph.D. degree in transport studies from the University of Leeds, Leeds, U.K., in 2015. In 2017, he was a Visiting Researcher with the Advanced Vehicle Engineering Centre, Cranfield University, Cranfield, U.K. He is currently an Associate Professor with the School of Mechanical Engineering, BIT. His research interests include intelligent transportation and vehicular systems, driver behavior modeling, reinforcement learning, and transfer learning and its applications.
\end{IEEEbiography}
\vspace{-10 mm}
\begin{IEEEbiography}[{\includegraphics[width=1in,height=1.25in,clip,keepaspectratio]{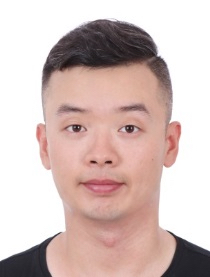}}]{Xinwei Wang}
is a Lecturer (Assistant Professor) at Queen Mary University of London (QMUL), UK. He was a Postdoc at TU Delft, The Netherlands from 2020 to 2022. Prior to that, he was a Postdoc at QMUL from 2019 to 2020, and he obtained a PhD degree from Beihang University, China in 2019. Over the years, he has integrated artificial intelligence and systems engineering for risk assessment, motion planning and decision making in intelligent systems. He is a recipient of Marie Sklodowska-Curie Actions Co-Fund Fellowship (2022), and IEEE ITSS Young Professionals Travelling Fellowship (2022). He has authored over 20 papers, including those in TR Part C, IEEE T-ITS, IEEE T-SMC, IEEE T-FS, IEEE T-VT, IEEE T-AES, etc.
\end{IEEEbiography}
\vspace{-10 mm}
\begin{IEEEbiography}[{\includegraphics[width=1in,height=1.25in,clip,keepaspectratio]{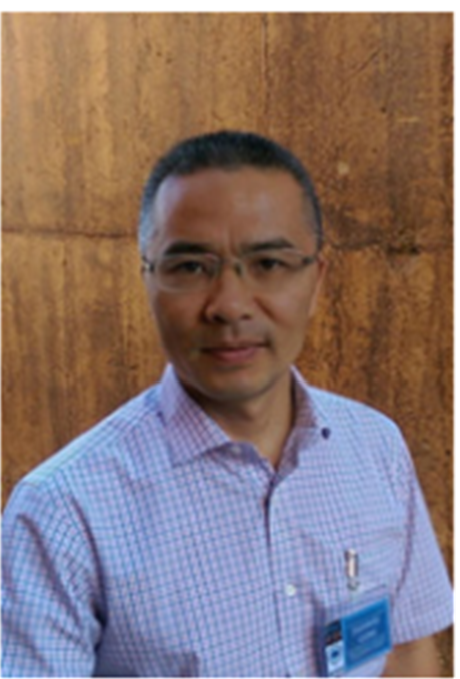}}]{Jianwei Gong}
received the B.S. degree from the National University of Défense Technology, Changsha, China, in 1992, and the Ph.D. degree from Beijing Institute of Technology, Beijing, China, in 2002. Between 2011 and 2012, he was a Visiting Scientist with the Robotic Mobility Group, Massachusetts Institute of Technology, Cambridge, MA, USA. He is currently a Professor with the School of Mechanical Engineering, Beijing Institute of Technology. His research interests include intelligent vehicle environment perception and understanding, decision making, path/motion planning, and control.
\end{IEEEbiography}
\end{document}